\documentclass[10pt,twocolumn,letterpaper]{article}

\usepackage{cvpr}      % To produce the REVIEW version
\usepackage{graphicx}
\usepackage{amsmath}
\usepackage{amssymb}
\usepackage{booktabs}
\usepackage{mathtools,bm,array}
\usepackage[accsupp]{axessibility}
\usepackage[symbol,hang,flushmargin]{footmisc}
\usepackage[pagebackref,breaklinks,colorlinks]{hyperref}

\usepackage[capitalize]{cleveref}
\crefname{section}{Sec.}{Secs.}
\Crefname{section}{Section}{Sections}
\Crefname{table}{Table}{Tables}
\crefname{table}{Tab.}{Tabs.}

\def\cvprPaperID{5884} % *** Enter the CVPR Paper ID here
\def\confName{CVPR}
\def\confYear{2023}

\begin{document}

\title{
 Robust Model-based Face Reconstruction 
through \\Weakly-Supervised Outlier Segmentation}

\author{ \normalsize{
Chunlu Li}$^{1,2}$
\quad
\normalsize{Andreas Morel-Forster} $^{2}$
\quad
\normalsize{Thomas Vetter} $^{2}$
\quad
\normalsize{Bernhard Egger}$^{3, *}$
\quad
\normalsize{Adam Kortylewski$^{4,5,*}$} \\
{\tt\small lcl@seu.edu.cn \quad \tt\small bernhard.egger@fau.de \quad \tt\small akortyle@mpi-inf.mpg.de}
\\
\normalsize{$^1$ School of Automation, Southeast University \quad $^2$ Department of Mathematics and Computer Science, University of Basel }\\ 
\normalsize{$^3$ Friedrich-Alexander-Universität Erlangen-Nürnberg \quad $^4$University of Freiburg \quad $^5$Max Planck Institute for Informatics  }
}

\maketitle
\footnotetext{$^*$ Denotes same contribution.}
\footnotetext{Codes available at: \href{https://github.com/unibas-gravis/Occlusion-Robust-MoFA}{github.com/unibas-gravis/Occlusion-Robust-MoFA}}
\footnotetext{C.Li is funded by the China Scholarship Council (CSC) from the Ministry of Education of P.R. China.
B.Egger was supported by a PostDoc Mobility Grant, Swiss National Science Foundation P400P2\_191110.
A.Kortylewski acknowledges support via his Emmy Noether Research Group funded by the German Science Foundation (DFG) under Grant No. 468670075.
Sincere gratitude to Tatsuro Koizumi and William A. P. Smith who offered the MoFA re-implementation.}
%%%%%%%%% ABSTRACT
\begin{abstract}
In this work, we aim to enhance model-based face reconstruction by avoiding fitting the model to outliers, i.e. regions that cannot be well-expressed by the model such as occluders or make-up.
The core challenge for localizing outliers is that they are highly variable and difficult to annotate.
To overcome this challenging problem, we introduce a joint Face-autoencoder and outlier segmentation approach (FOCUS).
In particular, we exploit the fact that the outliers cannot be fitted well by the face model and hence can be localized well given a high-quality model fitting.
The main challenge is that the model fitting and the outlier segmentation are mutually dependent on each other, and need to be inferred jointly.
We resolve this chicken-and-egg problem with an EM-type training strategy, where a face autoencoder is trained jointly with an outlier segmentation network.
This leads to a synergistic effect, in which the segmentation network prevents the face encoder from fitting to the outliers, enhancing the reconstruction quality.
The improved 3D face reconstruction, in turn, enables the segmentation network to better predict the outliers.
To resolve the ambiguity between outliers and regions that are difficult to fit, such as eyebrows, we build a statistical prior from synthetic data that measures the systematic bias in model fitting.
Experiments on the NoW testset demonstrate that FOCUS achieves SOTA 3D face reconstruction performance among all baselines trained without 3D annotation.
Moreover, our results on CelebA-HQ and AR database show that the segmentation network can localize occluders accurately despite being trained without any segmentation annotation.

\end{abstract}

\begin{figure}
  \centering
  \footnotesize
  \includegraphics[scale=0.23]{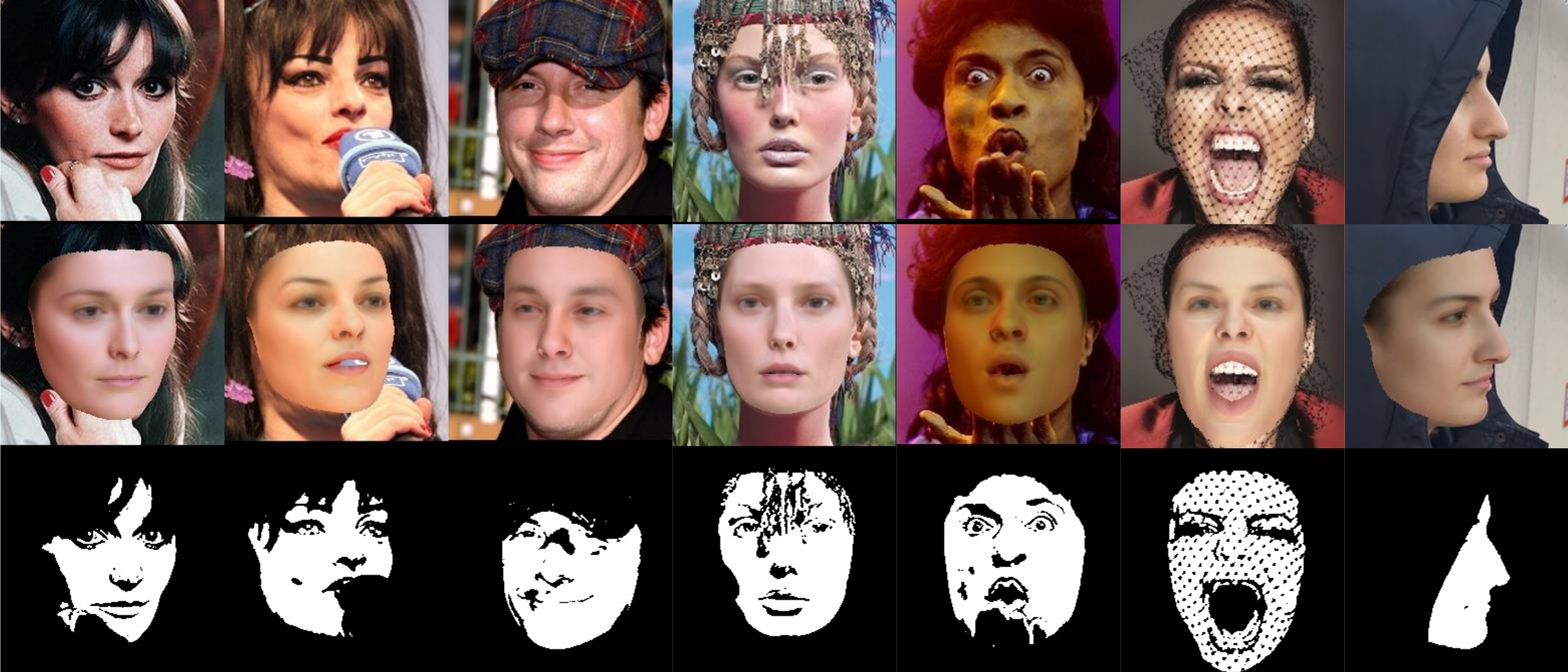}
  \caption{FOCUS conducts face reconstruction and outlier segmentation jointly under weak supervision. Top to bottom: target images, our reconstruction images, and estimated outlier masks.}
 \label{fig:intro}
\end{figure}

%%%%%%%%% BODY TEXT
\section{Introduction}
\label{sec:intro}

%------------------------------------------------------------
Monocular 3D face reconstruction aims at estimating the pose, shape, and albedo of a face, as well as the illumination conditions and camera parameters of the scene.
Solving for all these factors from a single image is an ill-posed problem.
Model-based face autoencoders \cite{tewari2017mofa} overcome this problem through fitting a 3D Morphable Model (3DMM) \cite{blanz2003face,egger20203d} to a target image.
The 3DMM provides prior knowledge about the face albedo and geometry such that 3D face reconstruction from a single image becomes feasible, enabling face autoencoders to set the current state-of-the-art in 3D face reconstruction \cite{deng2019accurate}. 
The network architectures in the face autoencoders are devised to enable end-to-end reconstruction and to enhance the speed compared to optimization-based alternatives \cite{kortylewski2018informed,zhu2015discriminative}, and sophisticated losses are designed to stabilize the training and to get better performance \cite{deng2019accurate}.

A major remaining challenge for face autoencoders is that their performance in in-the-wild environments is still limited by nuisance factors such as model outliers, extreme illumination, and poses. Among those nuisances, model outliers are ubiquitous and inherently difficult to handle because of their wide variety in shape, appearance, and location.
The outliers are a combination of the occlusions that do not belong to the face and the mismatches which are the facial parts but cannot be depicted well by the face model, such as pigmentation and makeup on the texture and wrinkles on the shape. Fitting to the outliers often distorts the prediction (see \cref{fig:OccusionAffectPerformance}) and fitting to the mismatches cannot improve the fitting further due to the limitation of the model. Therefore we propose to only fit the inliers, i.e. the target with the outliers excluded.

To prevent distortion caused by model outliers, existing methods often follow a bottom-up approach.
For example, a multi-view shape consistency loss is used as prior to regularize the shape variation of the same face in different images 
\cite{deng2019accurate,DECA,tiwari2022occlusion}, or the face symmetry is used to detect occluders \cite{8578512}. 
Training the face encoder with dense landmark supervision also imposes strong regularization \cite{wood20223d,zielonka2022towards}, while pairs of realistic images and meshes are costly to acquire.
Most existing methods apply face \cite{saito2016real} or skin \cite{deng2019accurate} segmentation models and subsequently exclude the non-facial regions during reconstruction. These segmentation methods operate in a supervised manner, which suffers from the high cost and efforts for acquiring a great variety of occlusion annotations from in-the-wild images.

In this work, we introduce an approach to handle outliers for model-based face reconstruction that is highly robust, without requiring any annotations for skin, occlusions, or dense landmarks.
In particular, we propose to train a Face-autOenCoder and oUtlier Segmentation network, abbreviated as FOCUS, in a cooperative manner.
To train the segmentation network in an unsupervised manner, we exploit the fact that the outliers cannot be well-expressed by the face model to guide the decision-making process of an outlier segmentation network.
Specifically, the discrepancy between the target image and the rendered face image (\cref{fig:intro} 1st and 2nd rows) are evaluated by several losses that can serve as a supervision signal
by preserving the similarities among the target image, the reconstructed image, and the reconstructed image under the estimated outlier mask.

The training process follows the core idea of the Expectation-Maximization (EM) algorithm, by alternating between training the face autoencoder given the currently estimated segmentation mask, and subsequently training the segmentation network based on the current face reconstruction.
The EM-like training strategy resolves the chicken-and-egg problem that the outlier segmentation and model fitting are dependent on each other.
This leads to a synergistic effect, in which the outlier segmentation first guides the face autoencoder to fit image regions that are easy to classify as face regions. The improved face fitting, in turn, enables the segmentation model to refine its prediction.

We define in-domain misfits as errors in regions, where a fixed model can explain but constantly not fit well, which are observed in the eyebrows and the lip region.
We assume that such misfits result from the deficiencies of the fitting pipeline.
Model-based face autoencoders use image-level losses only, which are highly non-convex and suffer from local optima.
Consequently, it is difficult to converge to the globally optimal solution.
In this work, we propose to measure and adjust the in-domain misfits with a statistical prior.
Our misfit prior learns from synthetic data at which regions these systematic errors occur on average.
Subsequently, the learnt prior can be used to counteract these errors for predictions on real data, especially when our FOCUS structure excludes the outliers.
Building the prior requires only data generated by a linear face model without any enhancement and no further improvement in landmark detection.

We demonstrate the effectiveness of our proposed pipeline by conducting experiments on the NoW testset \cite{Ringnet}, where we achieve state-of-the-art performance among model-based 3D face methods without 3D supervision.
Remarkably, experiments on the CelebA-HQ dataset \cite{CELEBAHQ} and the AR database \cite{ARdataset} validate that our method is able to predict accurate occlusion masks without requiring any supervision during training.

In summary, we make the following contributions:
\begin{enumerate}
\item We introduce an approach for model-based 3D face reconstruction that is highly robust, without requiring any human skin or occlusion annotation.
\item  We propose to compensate for the misfits with an in-domain statistical misfit prior, which is easy to implement and benefits the reconstruction.
\item  Our model achieves SOTA performance at self-supervised 3D face reconstruction and provides accurate estimates of the facial occlusion masks on in-the-wild images.
\end{enumerate}

\section{Related Work}

Model-based face autoencoders \cite{tewari2017mofa} solve the 3D face reconstruction by fitting a face model to the target image with an encoder and a decoder containing the 3DMM and a renderer. Typically, the encoder first estimates parameters from a target image, including the shape, texture, and pose of the target, and the illumination and camera settings from the scene. Then the renderer synthesizes a 2D image using the estimated parameters with an illumination model and a projection function. The face is reconstructed by retrieving the parameters which result in a synthesized image most similar to the target image. The 3DMM \cite{blanz2003face} plays a paramount role in the face autoencoders, because it parameterizes the latent distribution space of faces, and therefore can connect the encoder with the renderer and enable end-to-end training.
The model-based face autoencoders have been proven effective in improving the reconstruction. They simplify the optimization step and enhance the reconstruction speed \cite{Tewari_2018_CVPR}, improve the details of shape and texture \cite{DECA,Gecer_2019_CVPR,Richardson_2017_CVPR,8578512,tran2019towards}, and can also reconstruct more discriminative features \cite{deng2019accurate,genova2018unsupervised}.

Despite the advantages of face autoencoders, their performance is still limited by outliers.
To solve the distortions caused by outliers, some early methods \cite{romdhani2003efficient} resort to robust fitting losses, but they are not robust to illumination variations and appearance variations in eye and mouth regions. In recent years, shape consistency losses have been used as prior to constrain the face shape across images of the same subject \cite{deng2019accurate,DECA,Ringnet,tiwari2022occlusion}. The variation of identity features of the 3D shape is restricted so that the fitting remains robust even in unconstrained environments. However, such methods usually need identity labels and do not promise robust texture reconstruction. 
Besides, many methods conduct face segmentation before reconstruction to lead the model to avoid fitting outliers. For example, a random forest detector for hair is proposed \cite{morel2016generative} to avoid fitting the hair region, and a semantic segmentation network is trained to better locate the face region \cite{saito2016real}. A skin detector is employed to impose different weights on the pixels during reconstruction to guide the network to put more attention on the skin-colored regions \cite{deng2019accurate}. However, skin-colored occlusion can not be distinguished correctly and the detector is sensitive to illumination. Yildirim \etal propose to explicitly model the 3D shapes and cast shadows of certain types of occlusions, in order to decompose the target into face regions and occlusions to exclude occlusions during fitting \cite{yildirim2017causal}. However, the types of occlusions are limited.
Generally, these off-the-shelf segmentation models require labelled data for training. Although synthesized images can be used for training, there is a domain gap between the real images and the synthesized ones \cite{kortylewski2018training}. Unlike these methods, we merge the segmentation procedure into a model-based face autoencoder, which exploits the face model prior, and consequently does not require additional supervision.

The most relevant method to ours is proposed by Egger \etal. \cite{egger2018occlusion}. 
They jointly adapt a face model to a target image and segment the target image into face, beard, and occlusion, and the segmentation models are trained with an EM-like algorithm, where different models for beard, foreground, and background are optimized in alternating steps. Compared to our method, their model can only be inferred instance-wise, and the inferred likelihood models for one instance cannot be used for another, while ours follows the learning paradigm, which empowers the segmentation model with greater generalization ability and speeds up our inference procedure. Besides, building the beard model in \cite{egger2018occlusion} requires annotations for the beard, while our model is completely label-free. Additionally, their beard model and background likelihood model are statistical models based on simple color histogram, but our model is guided by perceptual-level losses measuring fitting quality, which are intuitive and much easier to implement.
De Smet \etal. also propose to conduct face model fitting and occlusion segmentation jointly \cite{1640924}, but they estimate the occlusions based on an appearance distribution model, which is sensitive to illumination variation and many other subtle changes in appearance. Maninchedda \etal. propose to solve face reconstruction and segmentation in a joint manner \cite{maninchedda2016semantic}, but depth maps are required as supervision. 
In comparison, our FOCUS model learns from only weak supervision.
The face autoencoder also enables us to adapt the face model more efficiently. In addition, we integrate perceptual losses which enable the segmentation network to reason over semantic features instead of only over independent pixels, which increases the robustness to illumination and other factors.

The in-domain misfits indicate systematic uncertainty and deficiencies in the fitting pipeline.
Instead of improving each part of the pipeline individually, we propose to build a statistical prior that measures and adjusts the bias introduced by the pipeline.
Our solution does not require any further improvements on the system such as landmark detection, model-landmark correspondence, or more supervision.

\begin{figure*}
  \centering
  \footnotesize
  \includegraphics[width=\textwidth]{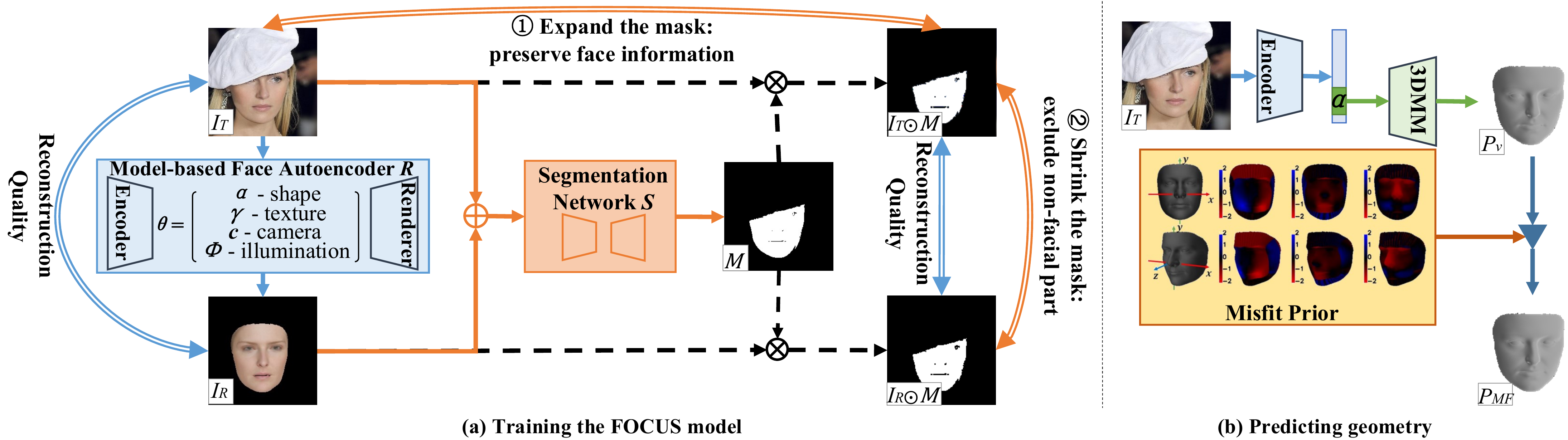}
  
  \caption{Overview of our method. The solid single lines show the forward path. (a) Given a target image $I_T$, the reconstruction network, $R$, estimates the latent parameters and subsequently renders an image $I_R$, containing only the face. Then, $I_T$ and $I_R$ are stacked and fed into the segmentation network, $S$, which predicts the mask $M$. The dashed lines show that $M$ is used to mask out the estimated outliers in $I_T$ and $I_R$ to get assembly outlier-free images, namely $I_T \odot M$ and $I_R\odot M$. The double-lined arrows indicate the compared image pairs in the losses for $S$ (orange) and losses for $R$ (blue), as stated in \cref{approach:EM}.
  By training alternatively the two networks and exploiting the synergy between the segmentation and the reconstruction tasks, the proposed FOCUS pipeline is capable of both reconstructing faces even under severe occlusions robustly and conducting face segmentation. 
  (b) Predicting the face geometry $P_{mf}$ requires a single forward. The $\blacktriangledown$ is a simple subtraction operation as introduced in \cref{approach:prior}.
}
 \label{fig:pipeline}
\end{figure*}

\section{Approach}

In this section, we introduce a neural network-based pipeline, FOCUS, that conducts 3D face reconstruction and outlier segmentation jointly.
We first discuss our proposed pipeline architecture (\cref{approach:Network}) and then the EM-type training without any supervision regarding the outliers (\cref{approach:EM}). In \cref{approach:initialization} we show the unsupervised EM initialization.
Finally, we show how to compensate for the systematic in-domain misfits with a statistical prior (\cref{approach:prior})

\subsection{Network Architecture}
\label{approach:Network}
Our goal is to robustly reconstruct the 3D face from a single target image $I_T$ with outliers, even severe occlusion.  
To solve this challenging problem, we integrate a model-based face autoencoder, $R$, with a segmentation network, $S$, and create synergy between them, as demonstrated in \cref{fig:pipeline}. For face reconstruction, the segmentation mask cuts the estimated outliers out during fitting, improving reconstruction robustness. For segmentation, the reconstructed result provides a reference, enhancing the segmentation accuracy. In this section, we explain how the two networks are connected together and how they benefit each other.

\textbf{The model-based face autoencoder}, $R$, is expected to reconstruct the complete face appearance from the visible face regions in the target image, $I_T$. It consists of an encoder and a computer graphics renderer as its decoder. The encoder estimates the latent parameters $\theta = [\alpha, \gamma,  \phi, c] \in \mathbb{R}^{257}$, i.e. the 3D shape $\alpha \in \mathbb{R}^{144} $ and texture $\gamma \in \mathbb{R}^{80} $ of a 3DMM, as well as the illumination $\phi  \in \mathbb{R}^{27} $ and camera parameters $c \in \mathbb{R}^{6}$ of the scene. 
Given the latent parameters, the decoder renders a predicted face image $I_R=\mathbb{R}(\theta)$.

\begin{figure}[t!]
  \centering
  \includegraphics[scale=0.23]{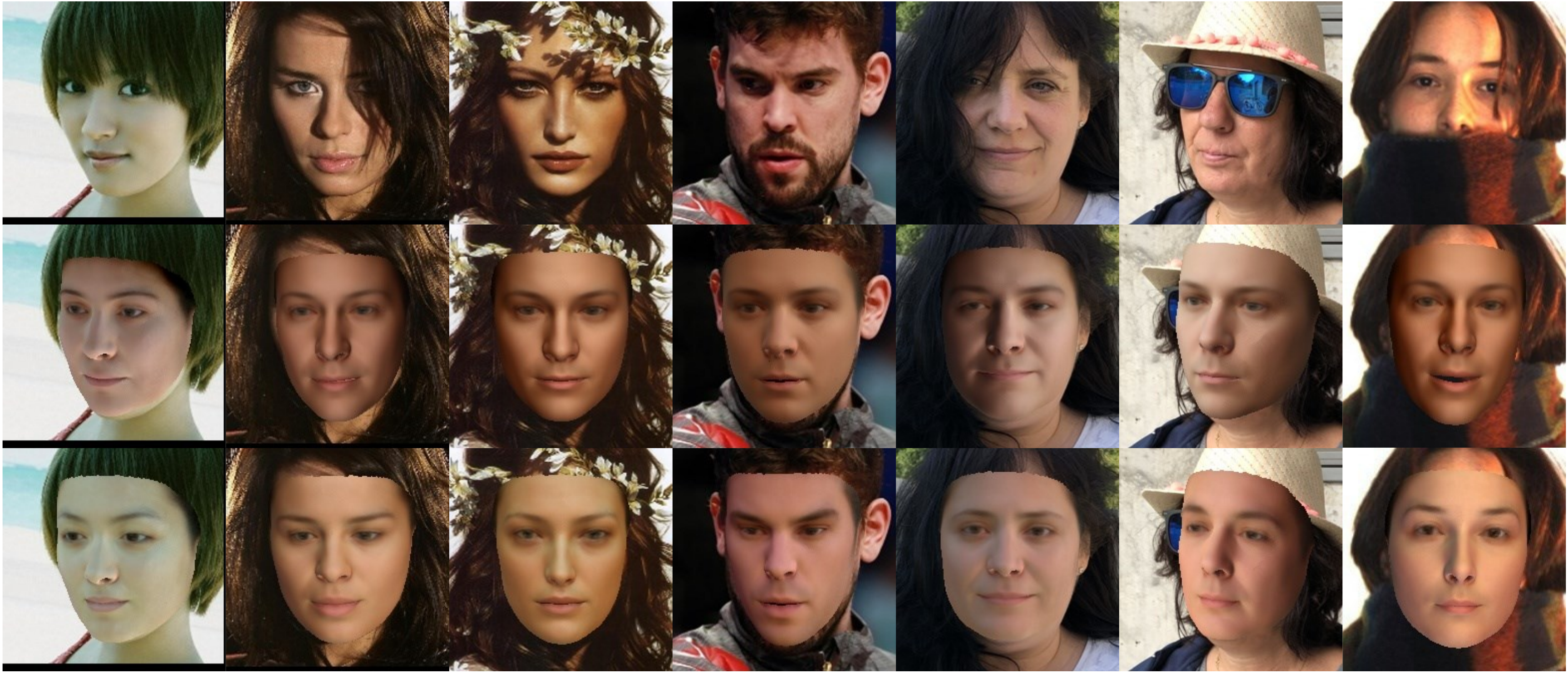}
  \caption{In the presence of outliers, FOCUS reconstructs faces more faithfully than previous model-based face autoencoders. The images from top to bottom are: target images, results of the MoFA network \cite{tewari2017mofa},
  and our results.}
 \label{fig:OccusionAffectPerformance}
\end{figure}

Standard face autoencoders \cite{tewari2017mofa} fit the face model parameters, regardless of whether the underlying pixels depict a face or occlusion. Consequently, the face model is distorted by the outliers, as shown in the second row in \cref{fig:OccusionAffectPerformance}, it is obvious that the illumination, appearance, and shape are estimated incorrectly.  
To resolve this fundamental problem of face autoencoders, we introduce an unsupervised segmentation network, whose output can be used to mask the outliers out during model fitting and therefore make the autoencoder robust to outliers.

\textbf{The segmentation network}, $S$, takes the stacked target image $I_T$ and the synthesized image $I_R$ as input and predicts a binary mask, $M=S(I_T,I_R)$, to describe whether a pixel depicts the face (1) or outliers (0). Since $I_R$ contains the estimated intact face, it provides the segmentation network with prior knowledge and helps the estimation.

The face autoencoder and the segmentation network are coupled together during training to induce a synergistic effect which makes the segmentation more accurate and reconstruction more robust under outliers, as shown in the last row in \cref{fig:OccusionAffectPerformance}.
\cref{approach:EM} describes how the pipeline can be trained end-to-end, despite the entanglement between the two networks, and the high-level losses that relieve our pipeline of any occlusion or skin annotation.

\subsection{EM-type Training}
\label{approach:EM}
Due to the mutual dependencies between the face autoencoder and the segmentation network, we conduct an Expectation-Maximization (EM) like strategy, where we train the two networks in an alternating manner. 
This enables a stable convergence of the model training process. Similar to other EM-type training strategies, our training process starts from a rough initialization of the model parameters which is obtained in an unsupervised manner (as described in \cref{approach:initialization}). We then optimize the two networks in an alternating manner, as described in the following.

\textbf{Training the segmentation network.} 
When training the segmentation network, the parameters of the face autoencoder are fixed and only the segmentation network is optimized. 
Instead of hunting for labelled data, we propose four losses enforcing intrinsic similarities among the images. Each loss works to either include pixels indicating face or the opposite. Since the proposed losses have overlapped or opposite functions to each other, only by reaching a balance among these losses can the network yield good segmentation results.
The losses work either on the perceptual level or the pixel level, to fully exploit the visual clues. The perceptual-level losses compare the intermediate features of two images extracted by a pretrained face recognition model $F$.
We use the cosine distance, $cos(X,Y)=1-{    \frac{X \cdot Y}{ \| {X} \|\| {Y} \|}}$, to compute the distance between the features. Perceptual losses are common for training face autoencoders, which encourage encoding facial details that are important for face recognition \cite{DECA}. We found that the perceptual losses also benefit segmentation (see \cref{ablation study}).

The proposed losses are as follows:
\begin{small}
\begin{gather}
  L_{nbr}= \sum_{x\in \Omega} \Big\| \min_{\forall x' \in \mathcal{N}(x)}  \big\| I_{T}(x) -  I_{R}(x') \big\| \Big\|_2^2 
  \label{eq:image exclude}
  \\
  L_{dist}= cos(F(I_{T}\odot M),F(I_{R}\odot M))
  \label{eq:perceptual exclude}
\\
  L_{area}= - S_M / S_R
    \label{eq:image include}
      \\
  L_{presv}= cos(F(I_{T}\odot M),F(I_{T}))
  \label{eq:perceptual include}
  \end{gather}
\end{small}
The pixel-level neighbor loss in \cref{eq:image exclude}, $L_{nbr}$, compares a pixel, $I_{T}(x)$, at location $x$ on the target image, with the pixels on the rendered image in the neighboring region, $\mathcal{N}(x)$ of this pixel, so that this loss is stable even if there are small misalignments. Note that it only accounts within the face region, $\Omega$, predicted by the segmentation network. A higher neighbor loss at $x$ indicates that this pixel is not fitted well and is more likely to be an outlier.

Similarly, $L_{dist}$ in \cref{eq:perceptual exclude} is introduced to compare the target and reconstructed face at perceptual level.
\cref{eq:image exclude,eq:perceptual exclude} aim at shrinking the mask on the outliers, where the pixel-level and perceptual differences are large. Without any other constraints, the segmentation network would output an all-zero mask to make them both 0. On the contrary, once there is a force to encourage the network to preserve some image parts, parts with smaller losses are more likely to be preserved, which in fact are the ones well-explained by the face model and is much more likely to depict face.

Therefore, \cref{eq:image include,eq:perceptual include} are proposed to counterwork \cref{eq:image exclude,eq:perceptual exclude}. \cref{eq:image include} is an area loss, $L_{area}$ that enlarges the ratio between the number of estimated facial pixels, $S_M$, and the number of pixels in the rendered face region, $S_R$. It prevents the segmentation network from discarding too many pixels. $L_{presv}$ (\cref{eq:perceptual include}), ensures that the perceptual face features remain similar after the outliers in the target image are masked out and encourages the model to preserve as much of the visible face region as possible. Likewise, the network would keep the most-likely face region to decrease \cref{eq:image include,eq:perceptual include} in the presence of \cref{eq:image exclude,eq:perceptual exclude}.

We use an additional regularization term,  $L_{bin}=-\sum_x ( M(x) -  0.5 )^2$, to encourage the face mask $M$ to be binary.
The total loss for training the segmentation network is:
{ $L_S= \eta_1 L_{nbr}+\eta_2 L_{dist}+\eta_3 L_{area}+\eta_4 L_{presv}+\eta_5 L_{bin}
$}, with $\eta_1=15$, $\eta_2 =3$, $\eta_3 =0.5$, and $\eta_4 =2.5$, and $\eta_5 =10$. Analysis of the influence of the hyper-parameters is provided in the supplementary material.

During training, the segmentation network is guided seeking a balance between discarding pixels that cannot be explained well by the face model and preserving pixels that are important to retain the perceptual representations of the target and rendered face images. Therefore no supervision for skin or occlusions is required.

\textbf{Training the face autoencoder.} In the second step, we continue to optimize the parameters of the face autoencoder, while keeping the segmentation network fixed. 
The losses for training the face autoencoder include:
    \begin{small}
   \begin{gather}
   L_{pixel}= \Big\|( I_{T} -  I_{R})\odot M\Big\|_2^2
    \label{eq:image reconstruction}
    \\
  L_{per}= cos(F(I_{T}),F(I_{R}))
  \label{eq:perceptual reconstruction}
  \\
    L_{lm}= \Big\| lm_T -  lm_R\Big\|_2^2
    \label{eq:landmarks}
    \end{gather}
\end{small}
Above are two reconstruction losses: $L_{pixel}$ (\cref{eq:image reconstruction}) at the image level and $L_{per}$ (\cref{eq:perceptual reconstruction}) at the perceptual level, and a landmark loss (\cref{eq:landmarks}) used to estimate the pose, where $lm_T$ and $lm_R$ stand for the 2D landmark coordinates on $I_T$ and $I_R$, respectively \cite{deng2019accurate}. We set the weights for the landmarks on the nose ridge and inner lip as 20, and the rest as 1. A regularization term is also required for the 3DMM: $L_{reg}= \Big\| \theta \Big\|_2^2$. To sum up, the loss for training the face autoencoder can be represented as: 
\begin{equation}
L_R= \lambda_1 L_{pixel} + \lambda_2 L_{per}+\lambda_3 L_{lm}+\lambda_4 L_{reg}
\label{eq:reconstruction loss sum}
\end{equation}, where $\lambda_1 =0.5$, $\lambda_2 =0.25$, $\lambda_3 =5e-4$, and $\lambda_4 =0.1$.

\subsection{Unsupervised Initialization}
\label{approach:initialization}
As every other EM-type training strategy, our training needs to be roughly initialized.
To achieve unsupervised initialization, we generate preliminary masks using an 
outlier robust loss \cite{egger2018occlusion}:
\begin{gather}
   log(P_{face}(x))=-\frac{1}{{2 {\sigma}^2 }} ( I_{T}(x) -  I_{R}(x))^2 + N_c
    \label{eq:P_face}
    \\
    M_{{pre}}(x)  =
    \begin{cases}
      1 & \text{if $( I_{T}(x) -  I_{R}(x) )^2< \xi$}\\
      0 & \text{otherwise}
    \end{cases}  
    \label{eq:pre-mask}
    \end{gather}
We assume that the pixel-wise error at pixel $x$ in the face regions follows a zero-mean Gaussian distribution. Therefore, we can express the log-likelihood that a pixel belongs to the face regions as $log(P_{face})$ (\cref{eq:P_face}), where $\sigma$ and $N_c$ are constant. We also assume that the values of the non-face pixels follow a uniform distribution, i.e., $log(P_{non-face})$ is a constant. Finally, a pixel at position $x$ is classified as face or non-face by comparing the log-likelihoods. This reduces to thresholding of the reconstruction error with a constant parameter $\xi$ (\cref{eq:pre-mask}). 
When $\xi$ increases, the initialized masks allow the pixels on the target image to have a larger difference from the reconstructed pixels and encourage the reconstruction network to fit these pixels. Empirically, we found that $\xi=0.17$ leads to a good enough initialization. 

To initialize the face autoencoder, the preliminary mask, $M_{{pre}}(x)$, is obtained in the forward pass using \cref{eq:pre-mask}, after the reconstructed face is rendered. Then $M_{{pre}}(x)$ is used to mask out the roughly-estimated outliers as in \cref{eq:image reconstruction}, preventing the face autoencoder from fitting to any possible outliers.
Subsequently, the segmentation network is pre-trained using these preliminary masks as ground truths.

\subsection{Solving misfits}
\label{approach:prior}
\begin{figure}
  \centering
  \includegraphics[scale=0.4]{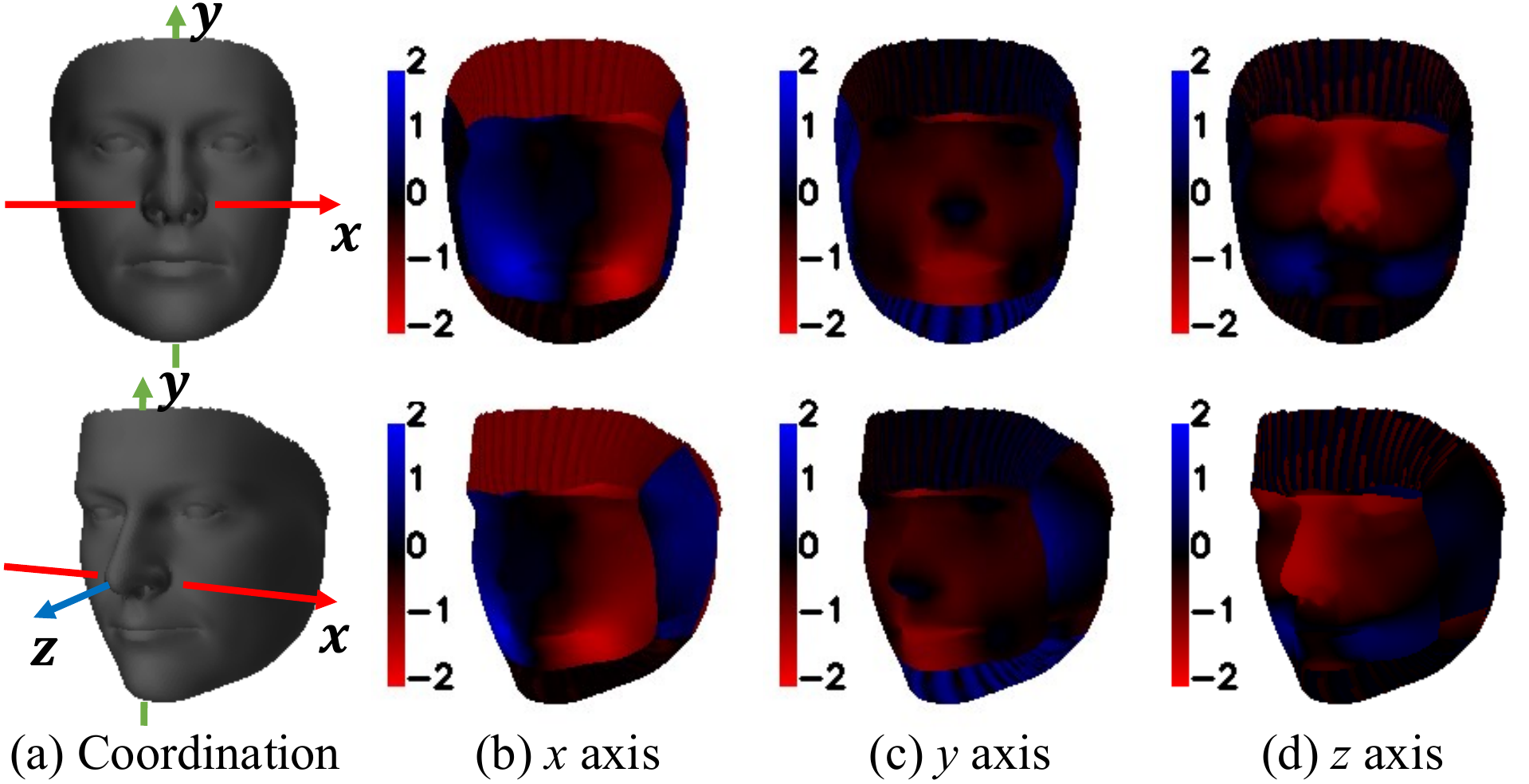}
  \caption{Visualization of the misfit prior. We provide two views of the coordinate system in (a) and the misfits along $x$, $y$, and $z$ axis in (b), (c), and (d), respectively.}
 \label{fig:misfit prior}
\end{figure}

The misfits in image regions that the model can explain yet not fitted well indicate systematic errors in the fitting pipeline. We propose an in-domain Misfit Prior (abbreviated as MP), $E_{MP}$, to measure and adjust such misfits.

To build the in-domain prior, we first synthesize images using the face model where theoretically every facial part should be fitted well. Hence,
any systematic error is due to systematic deficiencies. We draw random vectors $\theta_i$ in the face model latent space to generate ground truth (GT) $T_{vi}$ geometry and texture. Then the renderer in the face autoencoder is employed to render target images, $T_i$, $i=1,...,N$.

A face autoencoder, $R_{syn}$, with the same structure as $R$ is then trained on the synthesized images using losses \cref{eq:reconstruction loss sum}, except that no segmentation mask is required in pixel-wise loss, resulting in: $L_{pixel}= \Big\|I_{T} -  I_{R}\Big\|_2^2$, since there are no outliers on the synthesized images.
    
We built the statistical prior as the average error of the vertex-wise reconstruction deviation from the predicted geometry $P_{vi} \in \mathbb{R}^{px3}$ to the GT geometry $T_{vi}$, where $p$ is the number of vertices:
\begin{equation}
    E_{MP}= \frac{1}{N}\sum_{i\in [1,N]} (P_{vi}- T_{vi} )
    \label{eq:misfit}
\end{equation}
This prior visualized in \cref{fig:misfit prior} shows per-vertex bias introduced by the fitting pipeline. After inference, it could be used to adjust the in-domain misfits and the corrected prediction is $P_{mfi}= P_{vi} - E_{MP}$.

\section{Experiments}

In this section, results of systematic experiments show that our weakly-supervised method reaches the state-of-the-art face shape reconstruction accuracy and competitive occlusion segmentation results compared to the state-of-the-art self-supervised pipelines and methods that use full supervision in terms of skin or occlusion labels.
Our ablation study shows the effectiveness of the segmentation network, our proposed losses, and the misfit prior. More detailed analysis can be found in the supplementary material.
  \begin{figure}[b!]
  \centering
  
  \includegraphics[scale=0.23]{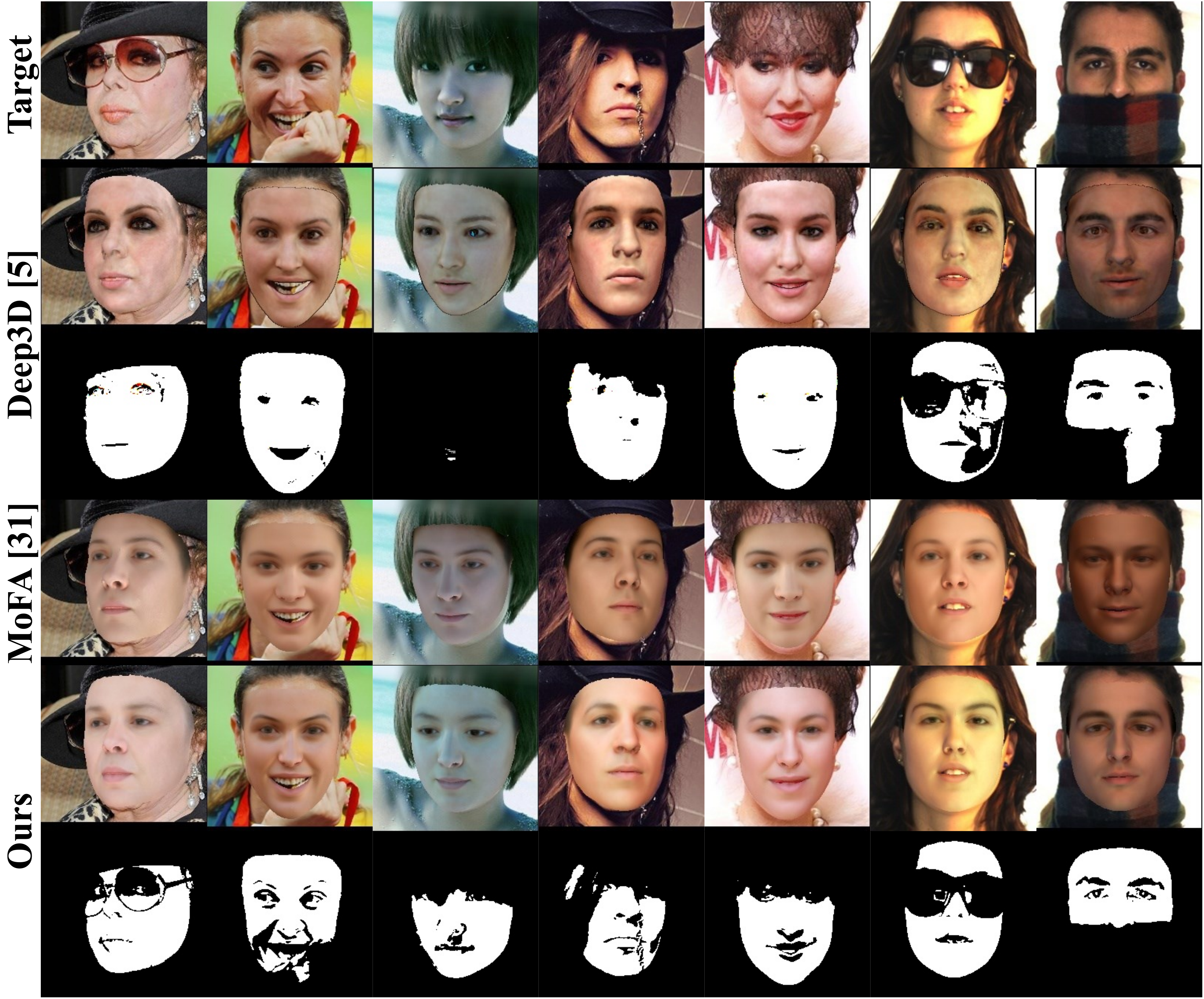}
  \caption{Qualitative comparison on the reconstruction and segmentation results of the Deep3D \cite{deng2019accurate} network (the 2nd and 3rd rows), the MoFA network \cite{tewari2017mofa} (the 4th row), and our FOCUS (the last two rows) on occluded faces from the CelebA-HQ testset (the first 8 columns) and the AR database (the last 2 columns). Note that all the masks are binarized.}
 \label{Visual Comparison}
\end{figure}
\subsection{Experiment setting}

\begin{table*}
  \caption{RMSE on the CelebA-HQ testsets and the AR testset.}
  \label{RMSE-fitting accuracy}
  \centering
 \footnotesize
  \renewcommand{\arraystretch}{0.95}
  \setlength{\abovecaptionskip}{1mm}
  \setlength{\belowcaptionskip}{1mm}
 {
  \begin{tabular}{c|cccccc}
  \hline
    Testset & MoFA \cite{tewari2017mofa}     & Backbone-Supervised  & Backbone-cutmix & Backbone-cutout & Deep3D \cite{deng2019accurate}  & FOCUS (ours)  \\
\hline
    CelebA-Unoccluded& $8.77 \pm 0.40$  & $8.71 \pm 0.38$ & $8.75 \pm 0.39$ & $8.72 \pm 0.40$ & $8.49 \pm 0.39$  & \bm{$8.38 \pm 0.42$}\\
    CelebA-Occluded  & $9.20 \pm 0.45$  & $9.01 \pm 0.45$ & $9.04 \pm 0.44$ & $8.99 \pm 0.45$ & $8.79 \pm 0.45$  & \bm{$8.71 \pm 0.48$}\\
    CelebA-Overall   & $8.99 \pm 0.47$  & $8.86 \pm 0.44$ & $8.90 \pm 0.44$ & $8.85 \pm 0.45$ & $8.64 \pm 0.44$  & \bm{$8.55 \pm 0.48$}\\
    AR-Overall       & $9.53 \pm 0.33$  & $9.34 \pm 0.33$ & $9.33 \pm 0.32$ & $9.28 \pm 0.31$ & $9.11 \pm 0.37$  & \bm{$8.93 \pm 0.35$}\\

  \end{tabular}
 }
\end{table*}

Our face encoder shares the structure of the ResNet 50 \cite{he2016deep} and uses the Basel Face Model (BFM) 2017 \cite{gerig2018morphable} as the 3D face model, with the differentiable renderer proposed in \cite{koizumi2020look}. The segmentation network follows the UNet architecture \cite{ronneberger2015u}. Our FOCUS pipeline is trained on the CelebA-HQ trainset \cite{CELEBAHQ}, following their protocol. Facial landmarks are detected using the method of \cite{bulat2017far}, and images are pre-processed in the same way as \cite{deng2019accurate}. The perceptual features are extracted by the pre-trained ArcFace \cite{deng2019arcface}.
More details can be found in the supplementary materials. Note that there is no fine-tuning with 3D data on any of the testsets in our experiments.

\textbf{Baselines.} We compare our method with two SOTA self-supervised model-based face autoencoders, i.e. the MoFA \cite{tewari2017mofa} and the Deep3D \cite{deng2019accurate}. Additionally, to achieve a fair comparison between our proposed weakly-supervised method and fully-supervised counterparts, we train our backbone reconstruction network in supervised settings using the GT masks provided by the CelebA-HQ database to exclude occlusions during training. The GT masks of the CelebA-HQ database are the merge of their manually labelled masks for skin, hair, accessories, and so on. Two data augmentation methods for occlusion handling, i.e. the cutmix \cite{yun2019cutmix} and cutout \cite{devries2017improved}, are also implemented to enhance the performance of the supervised pipelines. We refer to the three baselines as Backbone-Supervised, Backbone-cutmix, and Backbone-cutout, respectively. Note that we abbreviate a model with misfit prior as '-MP' 
for simplicity.

\textbf{Databases} We evaluate the shape reconstruction accuracy on the NoW database \cite{RingNet:CVPR:2019}. The publicly-available CelebA-HQ testset \cite{CELEBAHQ} and the AR database \cite{ARdataset} are used for validating the effectiveness of fitting and face segmentation. For the AR dataset, we take as GT masks the 120 manually-segmented masks in \cite{egger2018occlusion} that are publicly shared by the authors. The standard deviation is provided after $'\pm'$.

\subsection{Reconstruction Quality}

\cref{Visual Comparison} shows results of the Deep3D network, the MoFA network, and our method for qualitative comparison. Note that all the masks are binarized by rounding the pixels. The segmentation masks provided by the Deep3D result from a skin detector which assumes that skin color follows the simple multivariate Gaussian distribution. It shows that in our segmentation results, some small occlusions and skin-colored occlusions are better detected. Furthermore, our segmentation is more robust to illumination variations. It can also be observed from the reconstructed images that the illumination and texture of the faces are better estimated. Visually speaking, our method reaches competitive fitting results and improved outlier segmentation. More quantitative results are provided in the supplementary materials.

\textbf{Image fitting accuracy} shows how much the fitting results get misled by outliers. We evaluate the Root Mean Square Error (RMSE) between the input image and the reconstructed image inside visible face regions, with provided GT segmentation masks.
We compare different methods on the AR database, CelebA-HQ testset, and two randomly-selected occluded (750 images) and unoccluded subsets (558 images), referred to as 'CelebA-Overall', 'CelebA-Occluded', and 'CelebA-Unoccluded', respectively. As shown in \cref{RMSE-fitting accuracy}, our fitting accuracy is competitive even to the fully supervised counterpart with data augmentation.

\begin{table}
  \caption{Reconstruction error (mm) on the NoW testset \cite{RingNet:CVPR:2019}.} 
  \label{Now_Challenge}
  \centering

 \footnotesize
  \renewcommand{\arraystretch}{0.95}
  
  \setlength{\abovecaptionskip}{1mm}
  \setlength{\belowcaptionskip}{1mm}
  {

  \begin{tabular}{c|ccc}
     \hline
    Method&median& mean  & std \\
    \hline
    MICA \cite{zielonka2022towards} & \textbf{0.90} & \textbf{1.11} & \textbf{0.92} \\
    Wood \etal.\cite{wood20223d} & 1.02 & 1.28 & 1.08 \\

    DECA \cite{DECA}&1.09&1.38&1.18\\
    RingNet \cite{RingNet:CVPR:2019}&1.21&1.53&1.31\\
    Deep3D \cite{deng2019accurate}  &1.23&1.54&1.29 \\
    3DDFA V2 \cite{Guo:ECCV:2020} &1.23&1.57&1.39 \\
    Dib \etal. \cite{Dib:ICCV:2021} &1.26&1.57&1.31\\
    SynergyNet\cite{wu2021synergy} &1.27&1.59&1.31 \\
    MGCNet \cite{Shang:ECCV:2020} &1.31&1.87&2.63\\
    PRNet \cite{feng2018prn}&1.50&1.98&1.88\\
    3DMM-CNN \cite{tuan2017regressing} &1.84&2.33&2.05\\
    FOCUS (ours) &1.04&1.30&1.10\\
    FOCUS-MP (ours)  &  1.02&1.28&1.09\\
    \end{tabular}
  }

\end{table}

\begin{table}
  \caption{Reconstruction error (mm) on the non-occluded and occluded data in the NoW validation subset.} 
  \label{Now Subsets}
  \centering
  \footnotesize
  \renewcommand{\arraystretch}{0.95}
  \setlength{\abovecaptionskip}{1mm}
  \setlength{\belowcaptionskip}{1mm}{
  \tabcolsep=0.05cm

  \begin{tabular}{c|ccc|ccc}
     \hline
    & \multicolumn{3}{|c|}{Unoccluded Subset} & \multicolumn{3}{|c}{Occluded Subset}\\ 
    \hline
    Method  & median& mean  & std & median& mean & std  \\  
    \hline
    Deep3D \cite{deng2019accurate}  & 1.33&1.67&1.41&1.40&1.73&1.41\\
    DECA \cite{DECA} & 1.18 & 1.47 & 1.24 & 1.29 & 1.56 & 1.29\\ 
    MoFA \cite{tewari2017mofa}  &  1.35&1.69&1.42&1.36&1.69&1.41\\
    Backbone & 1.21& 1.46& 1.18& 1.33& 1.59& 1.27\\
    Backbone-Supervised & \textbf{1.02} & 1.25 & 1.04 & \textbf{1.05} & \textbf{1.29} & \textbf{1.09}\\
    Backbone-cutmix & 1.05 & 1.28 & 1.04 & 1.08 & 1.33 & 1.11\\
    Backbone-cutout & 1.03 & 1.28 & 1.06 & 1.09 & 1.34 & 1.10\\
    FOCUS (ours) &  1.03 & 1.25 & 1.03 & 1.07 & 1.34 & 1.19\\
    FOCUS-MP (ours) &  \textbf{1.02} & \textbf{1.24} & \textbf{1.02} & 1.08 & 1.34 & 1.20\\
  \end{tabular}
  }
\end{table}

\begin{table*}[t!]
  \caption{Evaluation of occlusions segmentation accuracy on the AR testsets.}
  \label{mask acc: AR}
  \centering
  \footnotesize
  \renewcommand{\arraystretch}{0.95}
  \setlength{\abovecaptionskip}{1mm}
  \setlength{\belowcaptionskip}{1mm}{
  
  \begin{tabular}{	c|cccc|cccc|cccc}
     \hline
    & \multicolumn{4}{|c|}{Unoccluded} & \multicolumn{4}{|c|}{Glasses} &\multicolumn{4}{|c}{Scarf}\\
    \hline

    Method  & ACC & PPV  & TPR& F1  & ACC & PPV & TPR& F1 & ACC & PPV & TPR & F1 \\
    \hline

    Deep3D \cite{deng2019accurate}  &      \textbf{0.88} &  0.93 &  \textbf{0.94} & \bm{$ 0.93 \pm 0.04$}  &
    \textbf{0.88} &  0.92 &  \textbf{0.92} & \bm{$ 0.92 \pm 0.04$}  & 0.80 &  0.80 &  \textbf{0.93} & $ 0.86 \pm 0.05$ \\
    Egger \etal. \cite{egger2018occlusion}&  -& -&- & 0.90   &- &- &- &0.87&  - & -&- & 0.86  \\
    FOCUS (ours) &  \textbf{0.88} &  \textbf{0.96} &  0.91 & \bm{$ 0.93 \pm 0.03$}  &
     \textbf{0.88} &  \textbf{0.98} &  0.85 & $ 0.91 \pm 0.04$  &
    \textbf{0.86} &  \textbf{0.97} &  0.81 & \bm{$ 0.88 \pm 0.05$}  \\
    
  \end{tabular}
  }
  \end{table*}

  \begin{table*}[t!]
  \caption{Ablation study on the AR testsets and the NoW validation subset.}
  \label{mask acc: Ablation}
  \centering
  \footnotesize
  \renewcommand{\arraystretch}{0.95}
  \setlength{\tabcolsep}{1.0mm}
  \setlength{\abovecaptionskip}{1mm}
  \setlength{\belowcaptionskip}{1mm}
  
  {
  \tabcolsep=0.1cm
  \begin{tabular}{	c|cccc|cccc|cccc|ccc}
     \hline
    & \multicolumn{4}{|c|}{AR-unoccluded} & \multicolumn{4}{|c|}{AR-glasses} &\multicolumn{4}{|c}{AR-scarf}&\multicolumn{3}{|c}{NoW Evaluation Set}\\
    \hline

    Method  & ACC & PPV  & TPR& F1  & ACC & PPV & TPR& F1 & ACC & PPV & TPR & F1 &median & mean & std \\
    \hline
    Pretrained & 0.75 &  0.95 &  0.77 & $ 0.85 \pm 0.05$  & 
    0.78 &  0.97 &  0.72 & $ 0.82 \pm 0.05$   &
    0.70 &  0.89 &  0.62 & $ 0.73 \pm 0.07$  &
    1.06&1.32&1.14\\
   
    Baseline   & 0.81 &  \textbf{0.96} &  0.83 & $ 0.89 \pm 0.04$   &
    0.81 &  0.97 &  0.76 & $ 0.85 \pm 0.05$  &
    0.79 &  0.96 &  0.71 & $ 0.82 \pm 0.07$  &
    1.06&1.32&1.15\\
    
    Neighbor  &0.85 &  0.95 &  0.88 & $ 0.91 \pm 0.04$ &
    0.84 &  0.95 &  0.81 & $ 0.87 \pm 0.04$ &
    0.83 &  0.94 &  0.79 & $ 0.85 \pm 0.06$ &
    1.06 & 1.32 & 1.15\\
    Perceptual & \textbf{0.89} &  \textbf{0.96} &  \textbf{0.92} & \bm{$ 0.94 \pm 0.03$}  &
    \textbf{0.89} &  \textbf{0.98} &  \textbf{0.87} & \bm{$ 0.92 \pm 0.04$}   & 
    \textbf{0.87} &  \textbf{0.97} &  \textbf{0.84} & \bm{$ 0.90 \pm 0.05$}  & 1.06&1.32&1.14 \\
    FOCUS   & 0.88 &  \textbf{0.96} &  0.91 & $ 0.93 \pm 0.03$  &
     0.88 &  \textbf{0.98} &  0.85 & $ 0.91 \pm 0.04$  &
    0.86 &  \textbf{0.97} &  0.81 & $ 0.88 \pm 0.05$ &
    1.05&1.31&1.14\\
    FOCUS-MF & -& -& -& -& -& -& -& -& -& -& -& -& \textbf{1.03} & \textbf{1.29} & \textbf{1.12}\\
  \end{tabular}
  }
  \end{table*}

\textbf{Shape reconstruction accuracy} is evaluated on the NoW Dataset. The cumulative errors on the testset in \cref{Now_Challenge} indicate that FOCUS reaches the state-of-the-art among the pipelines without 3D supervision even with a considerably smaller training set and no constraints on identity consistency. Note that \cite{zielonka2022towards,wood20223d} are trained with GT geometry of real images or synthesized images and we are comparable to the work of Wood \etal \cite{wood20223d}. To further evaluate the robustness fairly, 62 pairs of images in the evaluation set are selected with comparable poses with or without occlusions in the publicly-available validation set to normalize pose variation. \cref{Now Subsets} shows that the shape reconstruction accuracy of our pipeline is barely affected by occlusions, and reaches a similar level as the fully-supervised pipelines. Please refer to the supplementary for a more detailed analysis.

  \subsection{Outlier Segmentation}
  In this section, the performance of the outlier segmentation is indicated with occlusion segmentation accuracy, since occlusions account for a large portion of the outliers and their labels are feasible. Four indices are calculated inside the rendered regions, including accuracy (ACC), precision (Positive Predictive Value, PPV), recall rate (True Positive Rate, TPR), and F1 score (F1).
We separate the AR dataset into three subsets, which include faces without occlusions (neutral), faces with glasses (glasses), and faces with scarves (scarf).
According to \cref{mask acc: AR}, the masks predicted by our method show a higher accuracy, recall rate, and F1 score, and competitive precision compared to the skin detector used in \cite{deng2019accurate} and the segmentation method proposed in \cite{egger2018occlusion}. The results validate the potential of the outlier segmentation network to locate occlusions.

\subsection{Ablation Study}
\label{ablation study}

In this section, we first verify the utility of the segmentation network and the proposed neighbor loss, $L_{nbr}$ (\cref{eq:image exclude}), and the coupled perceptual losses, $L_{dist}$  (\cref{eq:perceptual exclude}) and $L_{presv}$ (\cref{eq:perceptual include}). We compare the segmentation performances of ablated pipelines on the AR testset, since it contains heavier occlusion, and test the shape reconstruction quality on the NoW evaluation subset. The pre-trained model as introduced in \cref{approach:initialization} is referred to as 'Pretrained'. We refer to the segmentation network trained without the neighbor loss or perceptual losses as 'Baseline' and \cref{eq:image reconstruction} is used to compensate for the lack of such losses. The 'Neighbor' and 'Perceptual' pipelines refer to the proposed segmentation network trained only with $L_{nbr}$ and only with two perceptual losses, $L_{dist}$ and $L_{presv}$, respectively.
The results in \cref{mask acc: Ablation} verify the usefulness of the segmentation network, since their results excel the pretrained model in almost all the indices. Comparison among the results of the ablated pipelines shows that both losses contribute significantly to the segmentation accuracy, indicating that the segmentations are more semantic with the FOCUS model. The gain in the reconstruction accuracy also validates the usefulness of the FOCUS model.
As for the misfit prior, results in \cref{Now_Challenge,Now Subsets,mask acc: Ablation} proves that it helps reduce the reconstruction error.
Please refer to the supplementary material for more comparisons among the ablated pipelines.

\subsection{Limitations}

Despite the accurate reconstruction and segmentation proven in the experiments, there are several limitations.
The main issue is that the potential of our outlier segmentation to occlusion segmentation is limited by the expressiveness of the face model. For further improvement, a model capable of depicting facial details, makeup, and beard is required.
Additionally, although the misfit prior reduces the overall reconstruction error, it does not promise enhancement for every single prediction.

\section{Conclusion}

In this paper, we address two sources of errors of the model-based face auto-encoder pipelines: the outliers and the misfits. We have shown how to solve face reconstruction and outlier segmentation jointly in a weakly-supervised way, so as to enhance the robustness for model-based face autoencoders in unconstrained environments. We have also shown how to reduce misfits with a statistical prior. Comprehensive experiments have shown that our method reaches state-of-the-art reconstruction accuracy on the NoW testset among methods without 3D supervision and provides promising segmentation masks as well.

Theoretically, our method can be integrated with the existing face autoencoders and/or non-linear parametric face models to achieve better performance. More importantly, we believe that the fundamental concepts of our approach can go beyond the context of face reconstruction and will inspire future work, such as human body reconstruction, or object reconstruction, with a reliable generative model.
We also expect that the masks will be useful for other tasks, \eg image completion, recognition, or more.
An analysis of the societal impact is provided in the supplementary materials.

%-------------------------------------------------------------------------

%%%%%%%%% REFERENCES
{\small
\bibliographystyle{ieee_fullname}
\bibliography{egbib}
}

%\title{Appendix}

\clearpage
\appendix

\setcounter{table}{0}

\setcounter{figure}{0}
\setcounter{page}{1}
In this section we reveal more implementation details and provide more analytical and visual comparisons on the shape reconstruction and segmentation performances.
\section{Implementation Details}

\subsection{The neighbor loss
}

\begin{figure}[h!]
  \centering
  
  \includegraphics[scale=0.33]{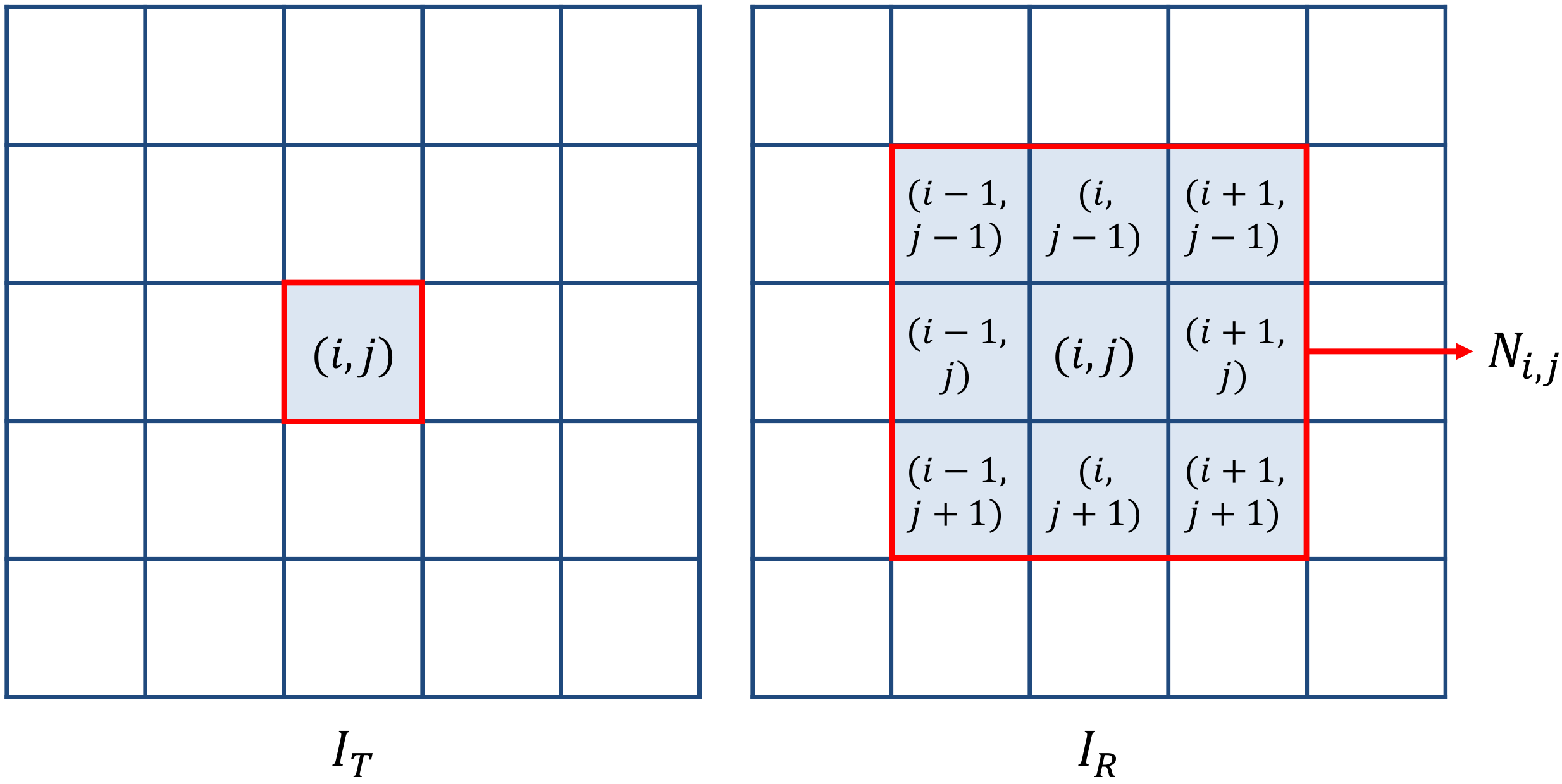}

    \caption{Visual explanation of the neighbor loss (\cref{eq:image exclude_supp}). For an pixel at $(i,j)$ on the target image $I_T$ (left), we compare it with its neighboring region, $N_{i,j}$, on the reconstructed image $I_R$ (right).}
 \label{Fig:Neighbor Loss}
\end{figure}

In this paper, we introduce a new image-level neighbor loss, $L_{nbr}$, that compares one pixel in the target image to a small region in the reconstructed image:
\begin{equation}
  L_{nbr}= \sum_{x\in \Omega}\Big\| \min_{\forall x' \in \mathcal{N}(x)}  \big\| I_{T}(x) -  I_{R}(x') \big\| \Big\|_2^2 
  \label{eq:image exclude_supp}
  \end{equation}
  
  As shown in \cref{Fig:Neighbor Loss}, for every pixel $I_{T_(i,j)}$ in the target image, we search in a $3 \times 3$ neighborhood $N{(i,j)}$ in the reconstructed image $I_{R}$  for the pixel that is most similar to $I_{T}(i,j)$ in intensity. This neighbour loss accounts for small misalignments of the face model during segmentation.

\subsection{Training details}

 To train our proposed pipeline, the Adadelta optimizer is used, with an initial learning rate of 0.1, and a decay rate of 0.99 at every 5k iterations. The learning rate for the segmentation network is 0.06 times the one for the reconstruction network. In every 30k iterations, 25k iters are for the face autoencoder training, and the rest are for training the segmentation network. For initialization, the face autoencoder is trained for 300k iterations. Afterwards, the face autoencoder and segmentation network are trained jointly for 200k iterations. The speed is evaluated on an RTX 2080 Ti, with batch size 12. 
It takes about 120 hours for the initialization of the face autoencoder, and about 80 hours to train the complete pipeline.
After the training, it takes 49 ms for reconstruction and 70 µs for segmentation on average for one image. The reconstruction
and the segmentation networks have 25.6M and 34.5M parameters,
respectively.

\section{Quantitative Analysis} 

\subsection{Reconstruction Performance.}

\begin{figure}
  \centering
  
  \includegraphics[scale=0.19]{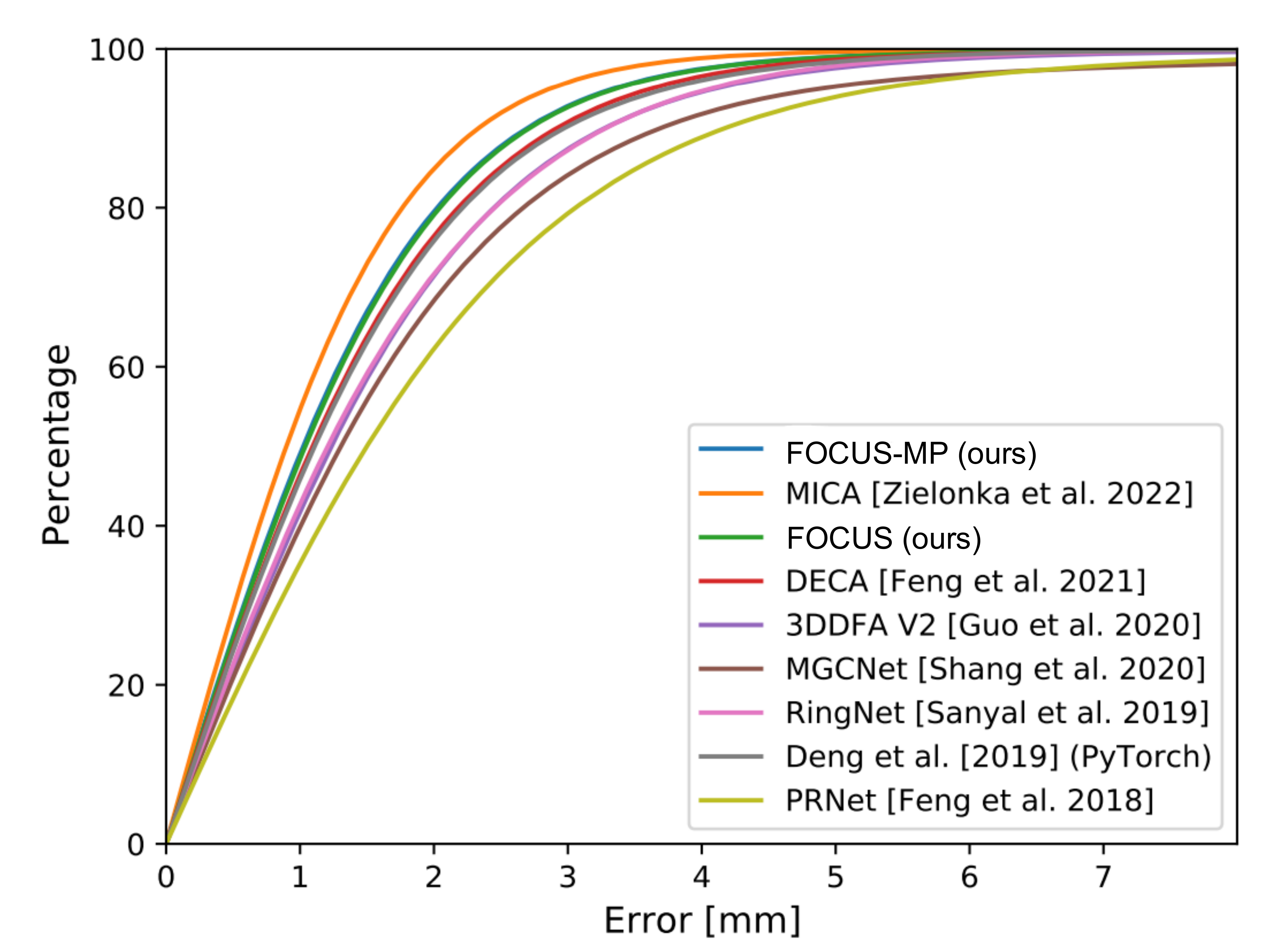}
  
    \caption{Quantitative comparison of the 3D reconstruction accuracy on the NoW \cite{RingNet:CVPR:2019} testset. The methods shown include: MICA \cite{zielonka2022towards}, DECA \cite{DECA}, the work of Dib et al. \cite{Dib:ICCV:2021}, 3DDFA V2 \cite{Guo:ECCV:2020}, MGCNet \cite{Shang:ECCV:2020}, RingNet \cite{RingNet:CVPR:2019}, Deep3D (pytorch version) \cite{deng2019accurate}, and PRNet \cite{feng2018prn}}
 \label{Analysis: NoW_AUC}
\end{figure}

\begin{figure*}
  \centering
  
  \includegraphics[scale=0.388]{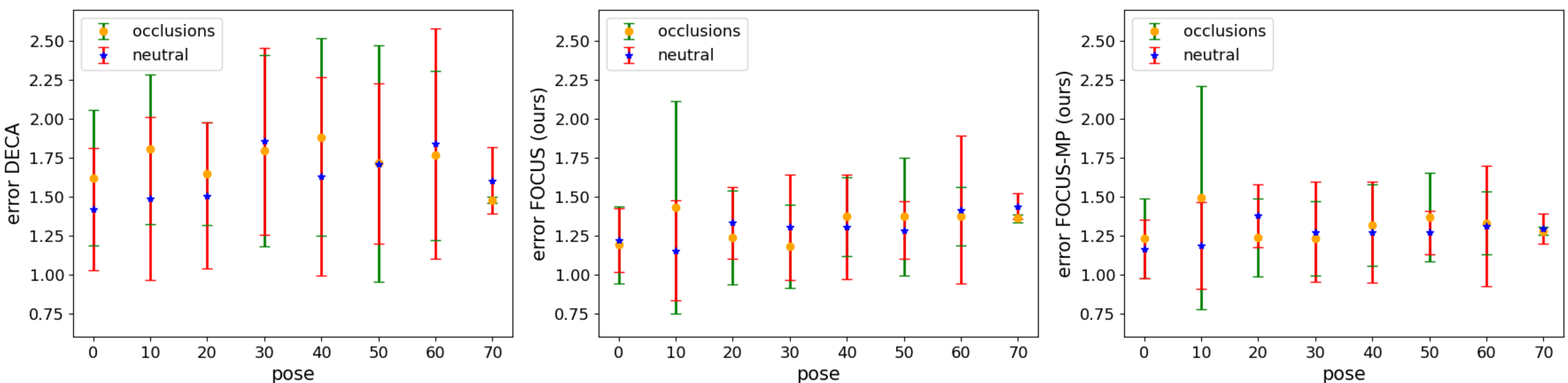}
  
    \caption{The distribution of the reconstruction errors on the neutral and occluded subsets of the full NoW validation set. The results of DECA \cite{DECA} are on the left, our FOCUS model in the middle and our FOCUS-MP on the right. The x axis indicates the approximated poses of the samples (rounded off to the nearest 10), and the y axis denotes the reconstruction error.}
 \label{Analysis: Occluded/unoccluded on Now}
\end{figure*}

\cref{Analysis: NoW_AUC} shows the cumulative error curves of the proposed method and the state-of-the-arts regarding the NoW testset. With a higher percentage of sampling points with lower errors, FOCUS performs the best on the NoW testset.

We further compare analytically the distributions of reconstruction errors of DECA \cite{DECA} and FOCUS on the NoW validation set, as shown in \cref{Analysis: Occluded/unoccluded on Now}. To further disentangle the influence of outliers from other factors, we categorize the samples according to the yaw angles (rounded off to the nearest 10), and use the error bars under different poses to reflect the distribution of the reconstruction errors. It is obvious from the plots that FOCUS exceeds DECA in mean errors and yields in much lower variations, even without identity supervision (which emphasize the shape consistency of a same identity) and with significantly less training data. Besides, the lower gap between errors and deviations under occluded and unoccluded conditions shows that the proposed method improves the outlier robustness. The comparison between the FOCUS and FOCUS-MP pipelines also indicates that the misfit prior improves the overall reconstruction accuracy.

\subsection{Segmentation Performance.}

\cref{mask acc: Celeb A} shows the segmentation performance on the Celeb A HQ testset, which indicates that the masks predicted by our method show a competitive accuracy, precision, and F1 score, compared to the skin detector used in \cite{deng2019accurate}.

 \begin{table*}
  \caption{Evaluation of occlusion segmentation accuracy on the CelebA-HQ testsets.}
  \label{mask acc: Celeb A}
  \centering
  \scalebox{1}{
  \tabcolsep=0.1cm
  \begin{tabular}{	c|cccc|cccc|cccc}
     \hline
    & \multicolumn{4}{|c|}{Unoccluded} & \multicolumn{4}{|c|}{Occluded} &\multicolumn{4}{|c}{Overall}\\
    \hline

    Method  & ACC & PPV  & TPR& F1  & ACC & PPV & TPR& F1 & ACC & PPV & TPR & F1 \\
    \hline

    Deep3D \cite{deng2019accurate}  &  \textbf{0.95} & 0.98 & \textbf{0.97} & \bm{$ 0.97 \pm 0.06$} & 0.84 & 0.86 & \textbf{0.96} &  $ 0.90 \pm 0.08$ & \textbf{0.89} & 0.92 & \textbf{0.96} &  \bm{$ 0.93 \pm 0.07$}   \\
    
    FOCUS (ours)                        &   0.92 & \textbf{0.99 } & 0.93 &  $0.96\pm 0.02$ & \textbf{0.86} & \textbf{0.95} & 0.87 & \bm{$0.91\pm 0.06$} & \textbf{0.89} & \textbf{0.97} & 0.90 & \bm{$0.93\pm0.05$}\\

  \end{tabular}
  }
  \end{table*}

 \subsection{Significance of the Misfit Prior}

Regarding Table 2 in the paper, a paired t-test on the NoW validation set shows a two-sided p-value of 3.4e-19, less than 0.05.
Hence, the mean errors before and after using the prior are not equal, indicating that the misfit prior brings a substantial increase in reconstruction accuracy.

\cref{Rebuttal-Misfit-Prior} shows the misfit prior generalizes well for our fully-supervised counterparts.

\begin{table}
  \caption{Reconstruction error (mm) on NoW validation subsets.} 
    \vspace{-.3cm}
  \label{Rebuttal-Misfit-Prior}
  \centering
  \scalebox{0.9}{
  \tabcolsep=0.01cm

  \begin{tabular}{c|ccc|ccc}
     \hline
    & \multicolumn{3}{|c|}{Unoccluded Subset} & \multicolumn{3}{|c}{Occluded Subset}\\ 
    \hline
    Method  & median& mean  & std & median& mean & std  \\  
    \hline
    
    Backbone-Supervised & 1.02 & 1.25 & 1.04 & 1.05 & 1.29 & 1.09\\
    Backbone-Supervised-MP & \textbf{1.00} &\textbf{1.23} &\textbf{1.02} &\textbf{1.03} &\textbf{1.28} &\textbf{1.08}\\
    \hline
    Backbone-cutmix & 1.05 & 1.28 & 1.04 & \textbf{1.08} & 1.33 & 1.11\\
    Backbone-cutmix-MP &\textbf{1.04} &\textbf{1.27} &\textbf{1.03} &\textbf{1.08} &\textbf{1.32} &\textbf{1.10} \\
    \hline
    Backbone-cutout & 1.03 & 1.28 & 1.06 & 1.09 & 1.34 & 1.10\\
    Backbone-cutout-MP &\textbf{1.02} & \textbf{1.25} & \textbf{1.04 }& \textbf{1.08} & \textbf{1.32} & \textbf{1.09}\\
  \end{tabular}
  }
\end{table}

\subsection{Hyper-parameter Analysis}

 In this section we systematically evaluate the influence of the hyper-parameters, $\eta_1$ to $\eta_5$, used for segmentation. The total loss for training the segmentation network is:
$L_S= \eta_1 L_{nbr}+\eta_2 L_{dist}+\eta_3 L_{area}+\eta_4 L_{presv}+\eta_5 L_{bin}$
, with $\eta_1=15$, $\eta_2 =3$, $\eta_3 =0.5$, and $\eta_4 =2.5$, and $\eta_5 =10$. We call this set of parameters as 'standard parameters'. We use the control variates method, namely changing one of the parameters while fixing the others, to compare the influence of each hyper-parameters. The accuracy (ACC), precision (Positive Predictive Value, PPV), recall rate (True Positive Rate, TPR), and F1 score (F1) reflect the segmentation performance. We use the AR dataset \cite{ARdataset} because the segmentation labels are more accurate.

\begin{figure*}
  \centering
  \includegraphics[scale=0.24]{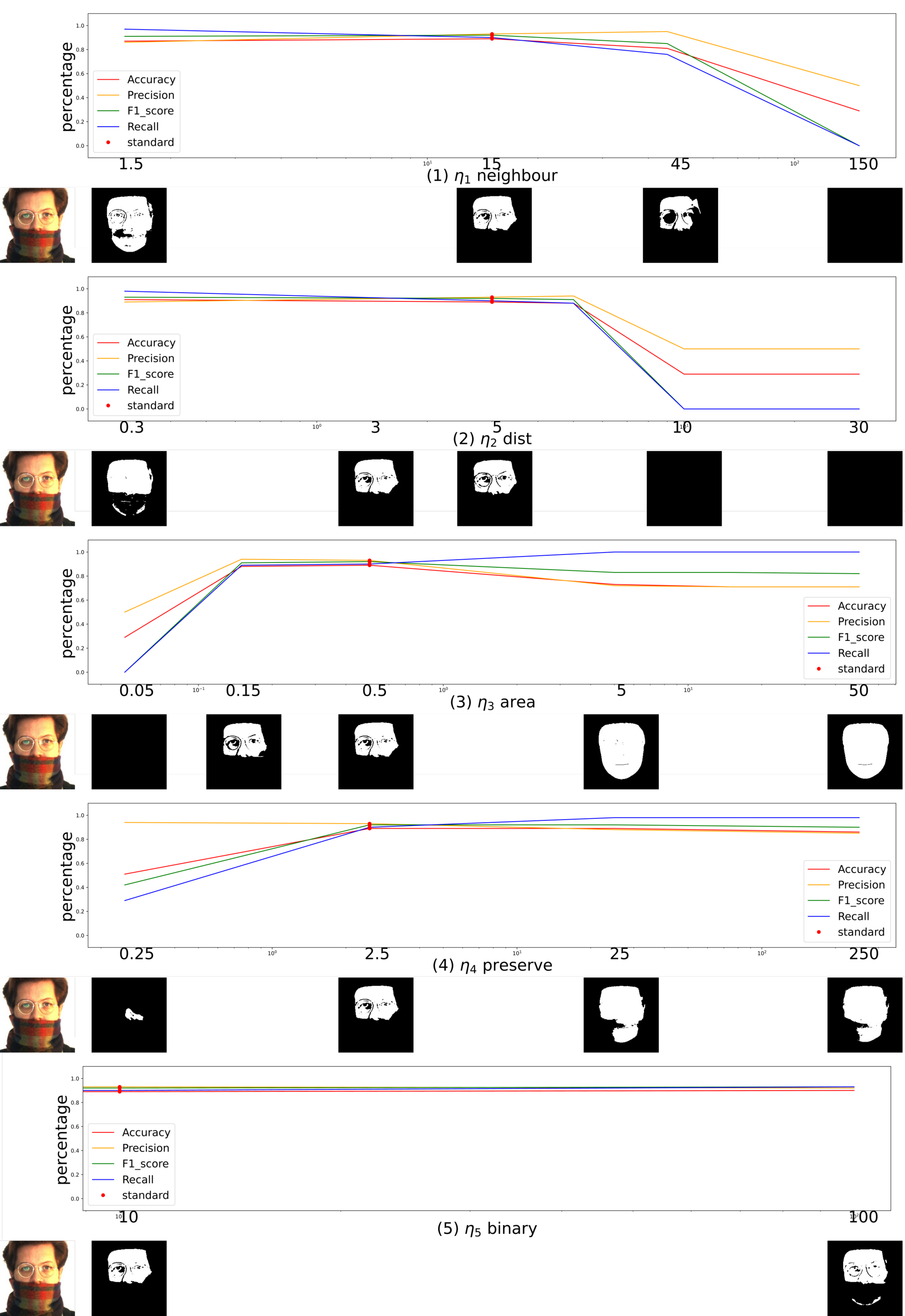}

    \caption{Analysis of hyper-parameters. The subplots show the change of for indices, namely accuracy, precision, F1 score, and recall rate, with the change of the hyper-parameters. The corresponding segmentation results are shown below each subplot. In each subplot, to evaluate the effect of each hyper parameter $\eta _i$, the other hyper-parameters $\eta_j(j\neq i)$ are fixed. The red dots denote the 'standard parameters' used in the paper.}
 \label{Analysis: Hyper-parameters}
\end{figure*}

As shown in \cref{Analysis: Hyper-parameters}, with the increase of the neighbour loss $L_{nbr}$ or the perceptual-level loss $L_{dist}$, more pixels are segmented as non-facial. On the contrary, when the area loss $L_{area}$ or the pixel-wise preserve loss $L_{presv}$ increases, more pixels are taken as face. This observation is consistent with our theory in section \textbf{3.2}. \cref{Analysis: Hyper-parameters} also indicates that the indices are positively related to the area loss $L_{area}$ and preserving loss $L_{presv}$, and are negatively related to the neighbour loss $L_{nbr}$ and the perceptual-level loss $L_{dist}$. The binary loss, $L_{bin}$, barely affects the segmentation.

Note that we did not excessively tune the loss weights, therefore we expect that better settings exist which achieve even higher performance.

 \section{Qualitative Comparison and Ablation Study}
In this section, we provide more visual results of out method on the Celeb A HQ testset \cite{CELEBAHQ}, the AR dataset \cite{ARdataset}, and the NoW testset \cite{RingNet:CVPR:2019}.

\begin{figure*}
  \centering  
  \includegraphics[scale=0.28]{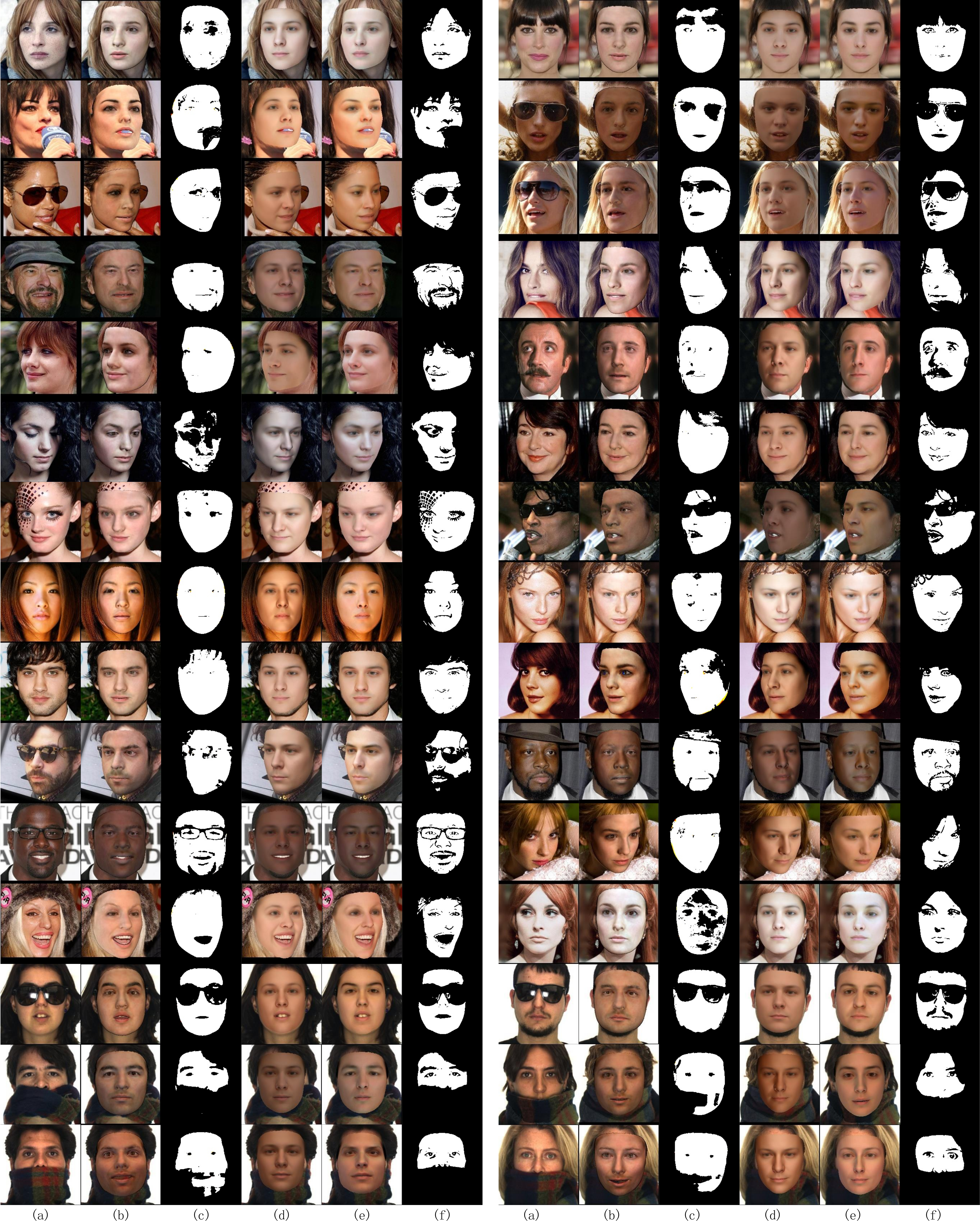}
  \caption{Comparison on \textbf{random samples} in the Celeb A HQ \cite{CELEBAHQ} testset and the AR dataset \cite{ARdataset}. (a) Target image. (b) and (c) Reconstruction and segmentation results of the Deep3D network \cite{deng2019accurate}. d) Reconstructed result of the MoFA network \cite{tewari2017mofa}. (e) and (f) Reconstruction and segmentation results of ours.}
 \label{Visual Comparison:general}
\end{figure*}

\begin{figure*}
  \centering

  \includegraphics[scale=0.28]{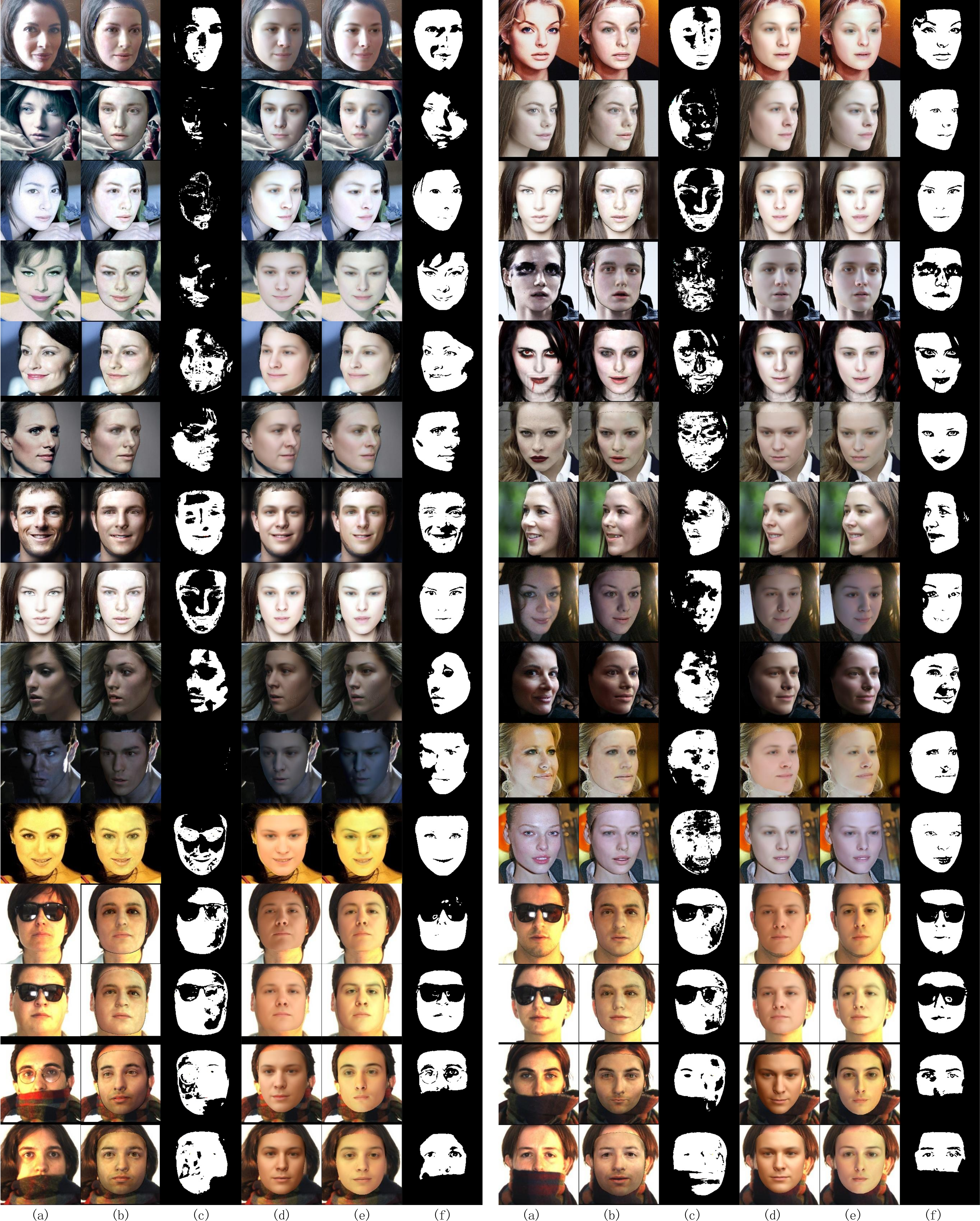}

    \caption{Comparison on samples with \textbf{extreme illumination} conditions in the Celeb A HQ \cite{CELEBAHQ} and the AR \cite{ARdataset} testsets. (a) Target image. (b) and (c) Reconstruction and segmentation results of the Deep3D network \cite{deng2019accurate}. d) Reconstructed result of the MoFA network \cite{tewari2017mofa}. (e) and (f) Reconstruction and segmentation results of ours.}
 \label{Visual Comparison:lighting}
\end{figure*}

\begin{figure*}
  \centering
  
  \includegraphics[scale=0.28]{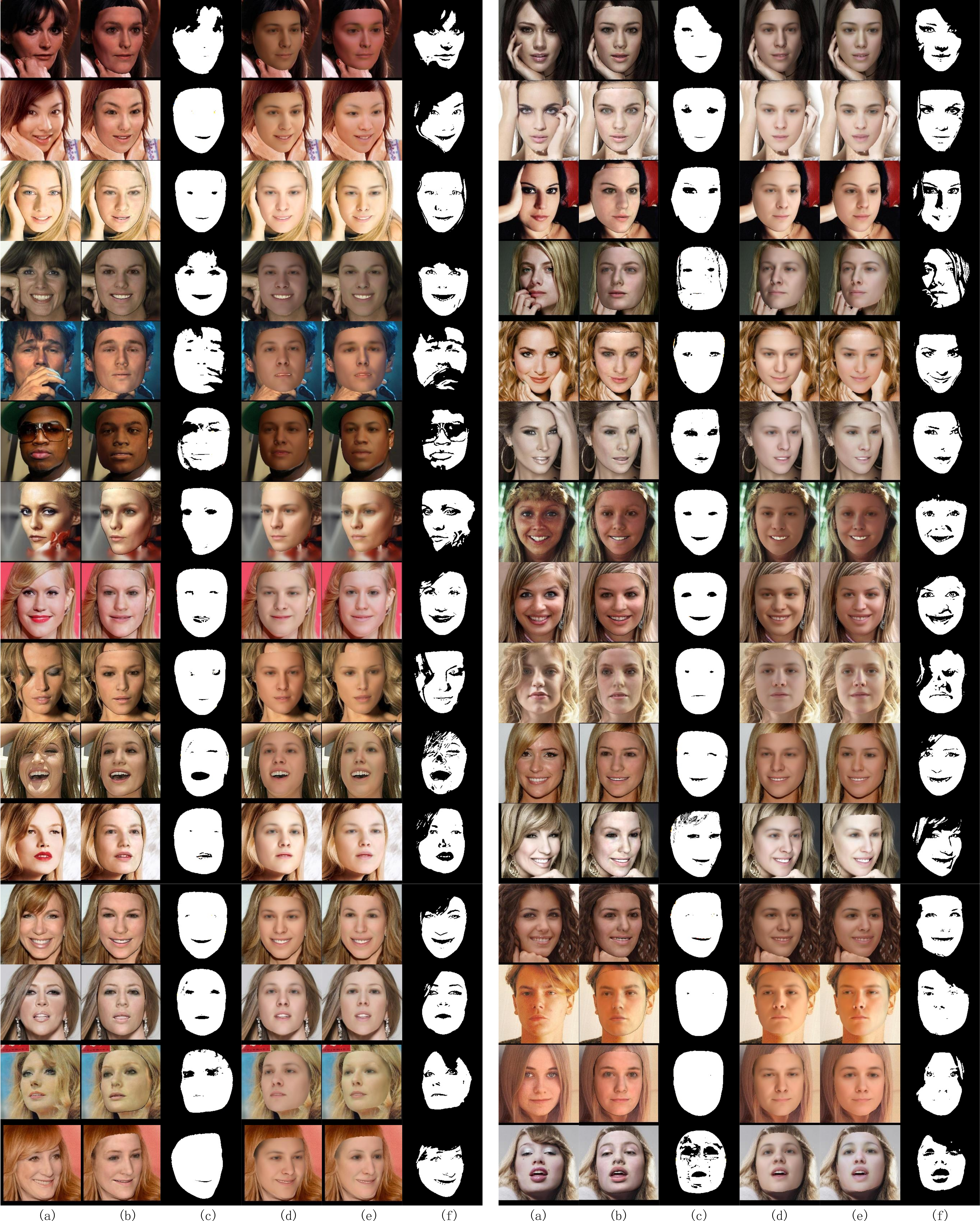}
  \caption{Comparison on samples with outliers that the \textbf{skin detector in \cite{deng2019accurate} fails} to locate in the Celeb A HQ testset \cite{CELEBAHQ}. (a) Target image. (b) and (c) Reconstruction and segmentation results of the Deep3D network \cite{deng2019accurate}. d) Reconstructed result of the MoFA network \cite{tewari2017mofa}. (e) and (f) Reconstruction and segmentation results of ours.}
  \label{Visual Comparison:skincolor}

\end{figure*}

\begin{figure*}
  \centering
  
  \includegraphics[scale=0.28]{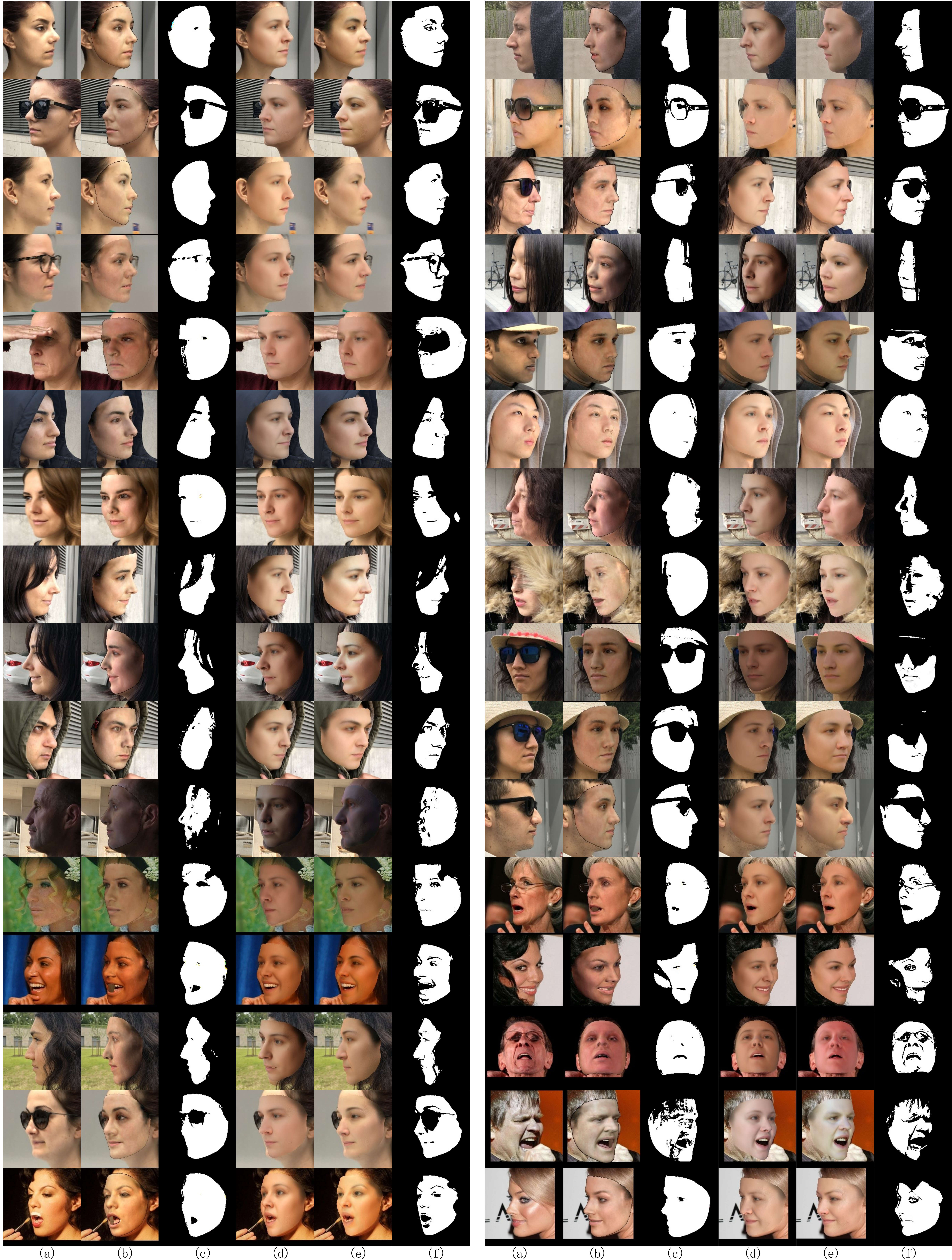}
  
  \caption{Comparison on samples with outliers and \textbf{large poses} in the NoW Database \cite{RingNet:CVPR:2019} shows that our method can effectively handle outliers even when there are large poses. (a) Target image. (b) and (c) Reconstruction and segmentation results of the Deep3D network \cite{deng2019accurate}. d) Reconstructed result of the MoFA network \cite{tewari2017mofa}. (e) and (f) Reconstruction and segmentation results of ours.}
  \label{Visual Comparison:poses}
\end{figure*}

\cref{Visual Comparison:general,Visual Comparison:lighting,Visual Comparison:skincolor,Visual Comparison:poses} show the results under general occlusions, extreme lighting, skin-colored occlusions, and large poses, respectively. 

\begin{figure}
\centering
  
  \includegraphics[scale=0.24]{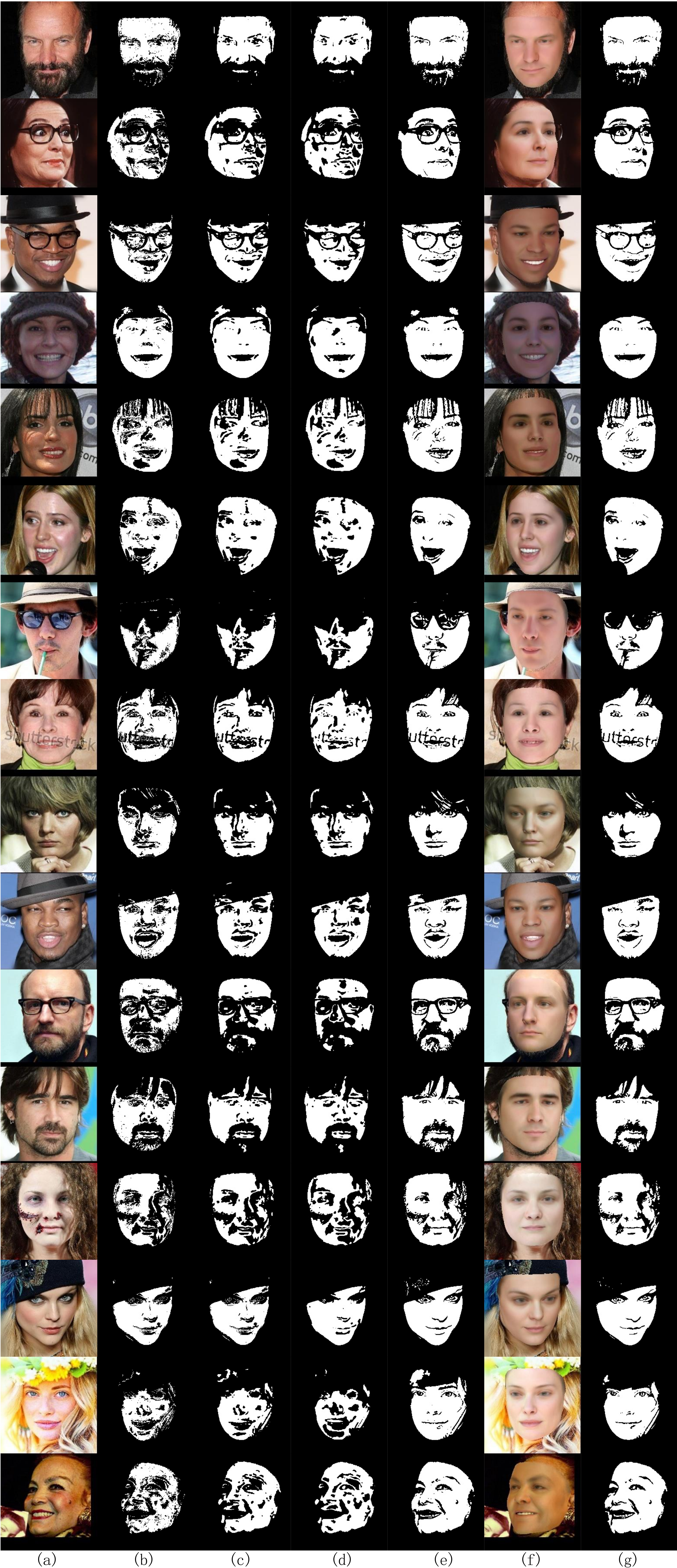}
  
  \caption{Qualitative comparison for ablation study on the Celeb A HQ testset \cite{CELEBAHQ}. From left to right are (a) target images, masks estimated by the (b) ’Pretrained’, (c) ’Baseline’, (d) ’Neighbour’, and (e) ’Perceptual’ pipelines, and (f) the reconstruction results and (g) predicted masks of FOCUS, respectively.}
  \label{Visual Comparison:ablation celeba}
\end{figure}

\begin{figure}
  \centering
  
  \includegraphics[scale=0.24]{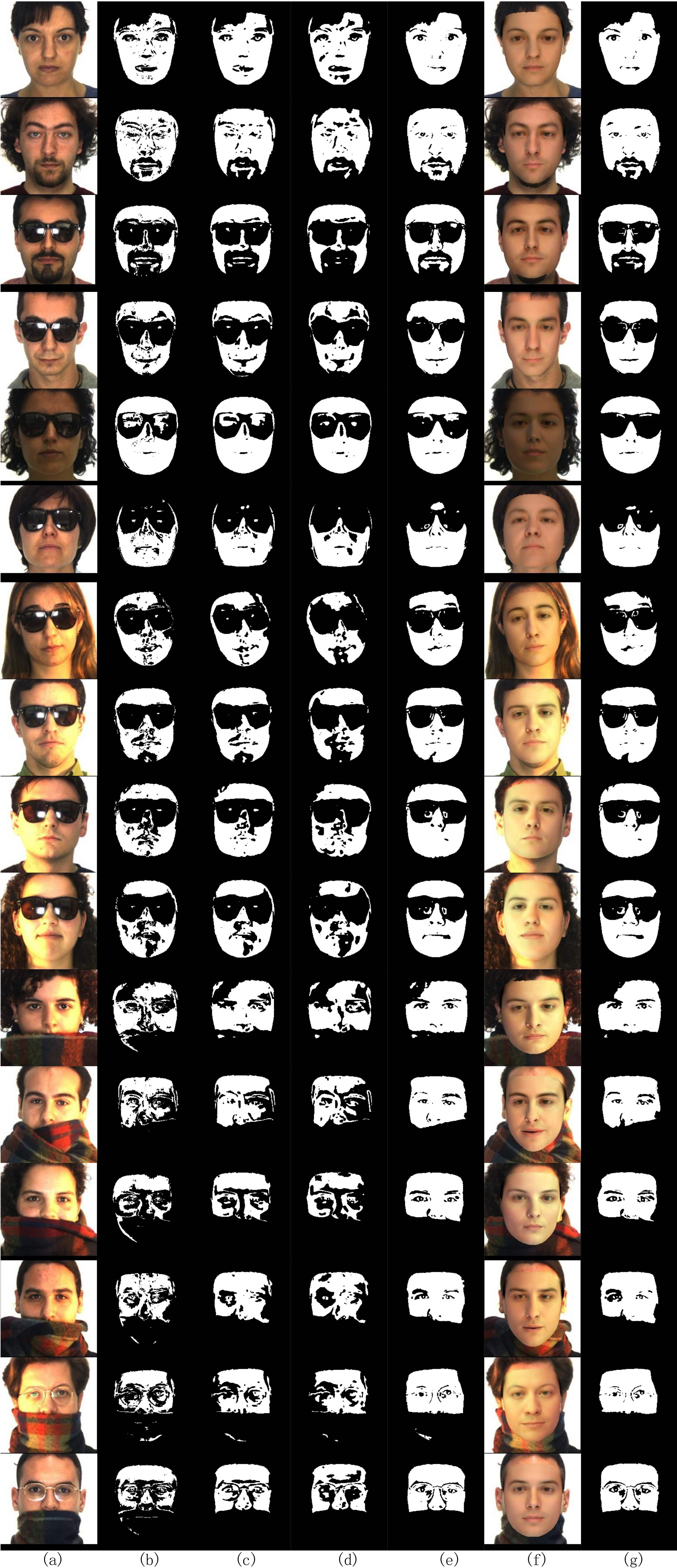}

  \caption{Qualitative comparison for ablation study on the AR testset \cite{ARdataset}. From left to right are (a) target images, masks estimated by the (b) ’Pretrained’, (c) ’Baseline’, (d) ’Neighbour’, and (e) ’Perceptual’ pipelines, and (f) the reconstruction results and (g) predicted masks of FOCUS, respectively.}
  \label{Visual Comparison:ablation AR}
\end{figure}

\cref{Visual Comparison:ablation celeba,Visual Comparison:ablation AR} provide a visual comparison among the ablated pipelines. It highlights that the outlier robust function is not robust to illumination variations, and the segmentation network brings great benefit to the robustness to illumination. The neighbour loss encourages the network to produce smoother results, and the perceptual losses help to locate the occlusions more accurately. Generally, the reconstruction performance of our proposed method are the best one and the segmentation accuracies are also competitive.

\begin{figure*}
  \centering
  
  \includegraphics[scale=0.32]{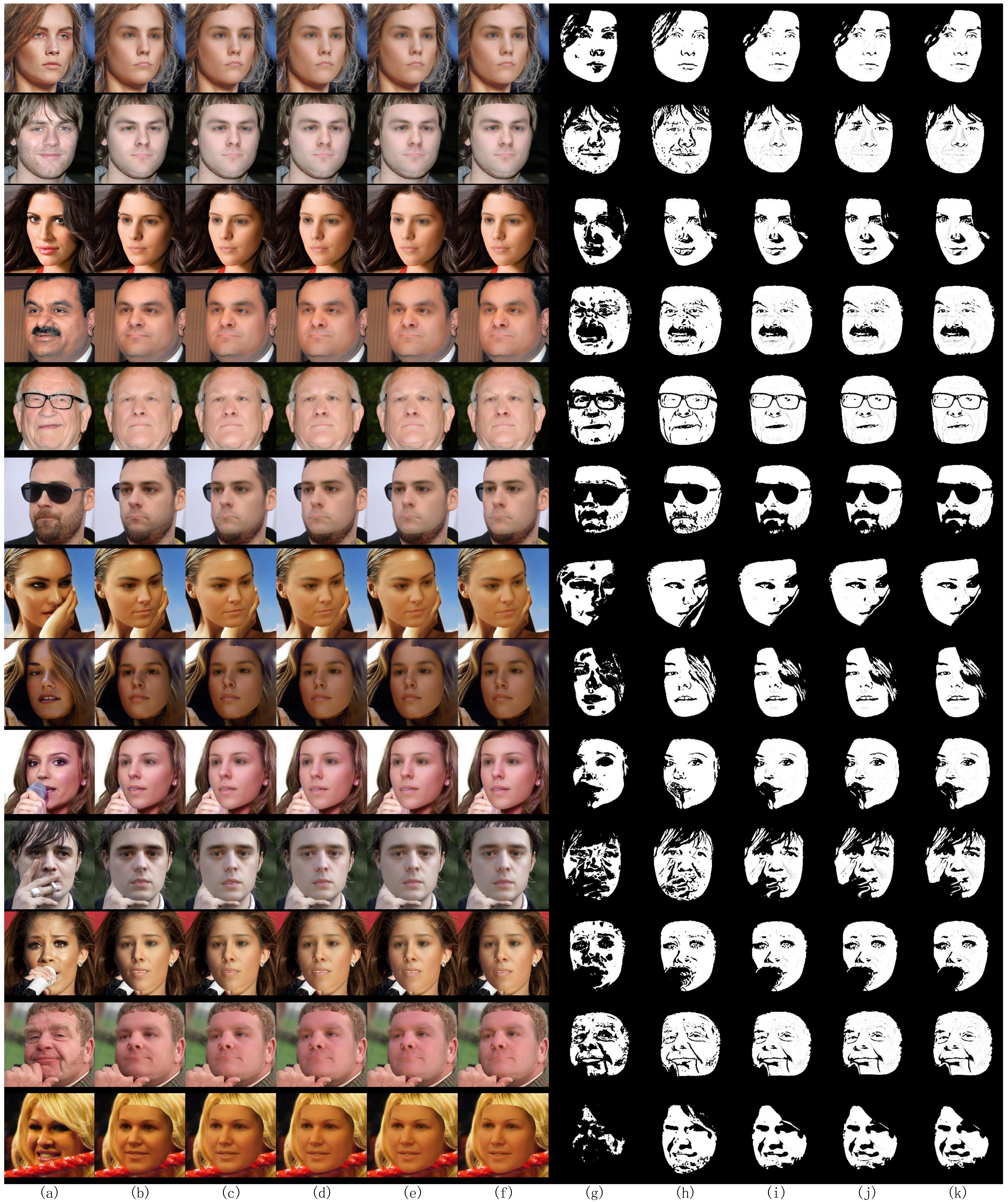}

  \caption{Target images (a) and intermediate results during the EM-like training. The intermediate masks and reconstructed faces predicted by: the initialized model introduced in section \textbf{3.3} (b and g), and the trained model after the first (c and h), second  (d and i), third (e and j), and last (f and k) round of EM training.}
  \label{Intermediate Results}
\end{figure*}

\cref{Intermediate Results} show the intermediate results during the EM-like training introduced in section \textbf{3.2}. 
The estimated masks get more accurate given better reconstruction results and the reconstructed faces show more details when provided with better segmentation masks, indicating the synergy effect of the reconstruction and segmentation networks.

\section{Societal Impact}
In general, our FOCUS pipeline has the potential to bring face reconstruction to the real world and to save costs of occlusion or skin labeling, which is generally required in many existing deep-learning-based methods.
The model-based reconstruction methods improved by our method could contribute to many applications, including Augmented Reality (AR), Virtual Reality (VR), surveillance, 3D design, and so on. Each of these applications may bring societal and economic benefits, and risks at the same time: The application of AR or VR could bring profits to the entertainment industry and also may result in unethical practices such as demonizing the image of others, identity fraud, and so on. The application of surveillance could help arrest criminals, yet might also invade the privacy and safety of others. The application in 3D design enables the quick capture of the 3D shape of an existing face but might also cause problems in portrait rights.

\end{document}